\documentclass{article} 
\usepackage{iclr2027_conference,times}

\iclrfinalcopy

\usepackage{amsmath,amsfonts,bm}

\def\eqref#1{equation~\ref{#1}}

\def\1{\bm{1}}

\DeclareMathAlphabet{\mathsfit}{\encodingdefault}{\sfdefault}{m}{sl}
\SetMathAlphabet{\mathsfit}{bold}{\encodingdefault}{\sfdefault}{bx}{n}

\usepackage{hyperref}
\usepackage{url}
\usepackage[utf8]{inputenc} 
\usepackage[T1]{fontenc}    
\usepackage{booktabs}       
\usepackage{amsfonts}       
\usepackage{nicefrac}       
\usepackage{microtype}      
\usepackage{xcolor}         
\usepackage{colortbl}
\usepackage{array}
\usepackage{amsmath,amssymb,amsfonts,mathtools}
\usepackage{multirow}
\usepackage{graphicx}
\usepackage{wrapfig}
\usepackage{natbib}
\usepackage{bbm}
\usepackage{amsmath}
\usepackage{tikz}
\usepackage{wrapfig}

\usetikzlibrary{arrows.meta,positioning,calc,fit,decorations.pathreplacing,backgrounds}

\definecolor{reasonblue}{RGB}{220,232,246}
\definecolor{promptgreen}{RGB}{228,237,228}
\definecolor{targetorange}{RGB}{247,224,180}
\definecolor{answergreen}{RGB}{221,240,221}
\definecolor{maskgray}{RGB}{238,238,238}
\definecolor{bordergray}{RGB}{190,190,190}
\definecolor{textgray}{RGB}{75,75,75}
\definecolor{wastepink}{RGB}{250,214,214}
\definecolor{losstext}{RGB}{153,51,0}
\definecolor{panelrule}{RGB}{160,168,180}
\definecolor{decisionlilac}{RGB}{230,224,244}
\definecolor{lmblue}{RGB}{203,221,241}
\definecolor{stepink}{RGB}{60,72,94}
\definecolor{oursrow}{RGB}{198,228,214}
\definecolor{headtint}{RGB}{238,242,246}
\definecolor{tablerule}{RGB}{168,176,186}
\definecolor{savedgreen}{RGB}{20,128,72}

\title{Learning to Stop without Learning to Stop: Self-Supervised Confidence Training Improves Reasoning Efficiency}

\author{%
  Parsa Hosseini$^{1,2}$\thanks{Correspondence to: \texttt{phoseini@umd.edu}} \quad 
   Akasha Tigalappanavara$^{2}$ \quad 
  Sumit Nawathe$^{1}$\quad
  Chenrui Fan$^{1}$\quad \\
\textbf{Sourya Basu$^{2}$ \quad  Genta Indra Winata$^{2}$ \quad Anirban Das$^{2}$ \quad 
  Soheil Feizi$^{1}$ \quad 
  Nima Chitsazan$^{2}$} \\
  $^{1}$University of Maryland \quad $^{2}$AI Foundations, Capital One
}

\begin{document}

\maketitle

\begin{abstract}
Reasoning models often generate very long reasoning traces, making inference computationally expensive. Existing approaches typically improve efficiency either through inference-time early-stopping mechanisms or by explicitly encouraging shorter reasoning during training, for example through reinforcement learning with length penalties. We show that substantial efficiency gains can instead emerge from a different kind of supervision: \textit{confidence}. Using a self-supervised procedure, we fine-tune reasoning models to predict their confidence in the answer at intermediate points along their own reasoning trajectories using only 600 training problems. Confidence is used only as a training target: the loss contains no objective for reasoning length, efficiency, or stopping. At inference, the fine-tuned models use the standard generation procedure, with no confidence elicitation or early-stopping mechanism. Despite this, self-supervised confidence fine-tuning makes reasoning more efficient, reducing generated tokens by up to 25\% at matched accuracy across Gemma, Qwen, Nemotron, and GPT-OSS models on mathematical, scientific, and coding reasoning benchmarks, with efficiency gains comparable to methods that explicitly optimize for shorter reasoning. Analysis of reasoning episodes further shows that confidence supervision largely preserves the base models' high-level reasoning composition rather than selectively suppressing particular behaviors. Our results suggest that efficient reasoning may emerge as a downstream consequence of learning metacognitive signals, without being directly optimized.

\end{abstract}

\section{Introduction}

Reasoning language models have achieved strong performance across a wide range of challenging tasks, but often at substantial inference cost. Modern reasoning models can generate tens of thousands of tokens before producing an answer, motivating growing interest in making reasoning more efficient.

\definecolor{softblue}{RGB}{247,249,255}
\definecolor{lineblue}{RGB}{72,88,190}
\definecolor{softgreen}{RGB}{239,250,241}
\definecolor{linegreen}{RGB}{50,145,70}
\definecolor{panelgray}{RGB}{250,250,250}
\definecolor{textgray}{RGB}{105,105,105}

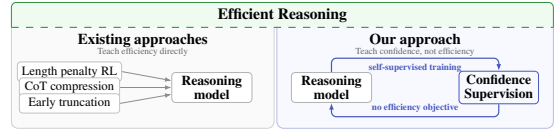
\begin{wrapfigure}{r}{0.52\linewidth}
\vspace{-0.6\baselineskip}
\centering

\resizebox{\linewidth}{!}{%
\begin{tikzpicture}[
 x=1cm,y=1cm,>=Latex,
 method/.style={draw=black!24,rounded corners=1.4pt,fill=white,minimum width=1.45cm,minimum height=.28cm,align=center,font=\fontsize{5.2}{5.6}\selectfont,inner xsep=1.8pt,inner ysep=.6pt},
 model/.style={draw=black!30,rounded corners=1.6pt,fill=white,minimum width=1.16cm,minimum height=.42cm,align=center,font=\fontsize{5.6}{6.0}\selectfont\bfseries,inner xsep=2pt,inner ysep=1.2pt},
 conf/.style={draw=lineblue,rounded corners=1.6pt,fill=softblue,minimum width=1.15cm,minimum height=.42cm,align=center,font=\fontsize{5.6}{6.0}\selectfont\bfseries,inner xsep=2pt,inner ysep=1.2pt},
 grayarr/.style={-{Latex[length=.9mm]},line width=.36pt,draw=black!45},
 bluearr/.style={-{Latex[length=.95mm]},line width=.50pt,draw=lineblue,rounded corners=2pt},
 title/.style={font=\fontsize{6.2}{6.6}\selectfont\bfseries},
 sub/.style={font=\fontsize{3.8}{4.2}\selectfont,text=textgray},
 loopmain/.style={font=\fontsize{3.8}{4.1}\selectfont\bfseries,text=lineblue,align=center},
 loopsub/.style={font=\fontsize{3.8}{4.1}\selectfont\bfseries,text=lineblue,align=center}
]

\def\Lx{0.00}
\def\Lr{3.80}
\def\Rx{3.84}
\def\Rr{7.84}
\def\PyTop{1.46}
\def\PyBot{0.00}
\def\HyBot{1.46}
\def\HyTop{1.82}


\fill[panelgray]
  (\Lx,\PyBot) rectangle (\Lr,\PyTop);

\fill[softblue]
  (\Rx,\PyBot) rectangle (\Rr,\PyTop);

\fill[softgreen]
  (\Lx,\HyBot) rectangle (\Rr,\HyTop);


\draw[black!18,rounded corners=2.8pt]
  (\Lx,\PyTop)
  -- (\Lx,\PyBot)
  -- (\Lr,\PyBot)
  -- (\Lr,\PyTop);

\draw[lineblue!40,rounded corners=2.8pt]
  (\Rx,\PyTop)
  -- (\Rx,\PyBot)
  -- (\Rr,\PyBot)
  -- (\Rr,\PyTop);


\draw[linegreen,rounded corners=2pt]
  (\Lx,\HyBot)
  -- (\Lx,\HyTop)
  -- (\Rr,\HyTop)
  -- (\Rr,\HyBot);

\draw[
  linegreen,
  dashed,
  line width=.45pt
]
  (\Lx,\HyBot) -- (\Rr,\HyBot);

\node[
  font=\fontsize{6.4}{6.8}\selectfont\bfseries
] at (3.92,1.64)
  {Efficient Reasoning};


\node[title] at (1.90,1.27)
  {Existing approaches};

\node[sub] at (1.90,1.11)
  {Teach efficiency directly};

\node[title] at (5.84,1.27)
  {Our approach};

\node[sub] at (5.84,1.11)
  {Teach confidence, not efficiency};


\node[method] (rl) at (.84,.82)
  {Length penalty RL};

\node[method] (cot) at (.84,.59)
  {CoT compression};

\node[method] (tr) at (.84,.36)
  {Early truncation};

\node[model] (mold) at (2.90,.59)
  {Reasoning\\model};

\draw[grayarr]
  (rl.east) -- ($(mold.west)+(0,.10)$);

\draw[grayarr]
  (cot.east) -- (mold.west);

\draw[grayarr]
  (tr.east) -- ($(mold.west)+(0,-.10)$);


\node[model] (mour) at (4.63,.59)
  {Reasoning\\model};

\node[conf] (cf) at (7.05,.59)
  {Confidence\\Supervision};

\draw[bluearr]
  (mour.north)
  -- ++(0,.20)
  -| (cf.north);

\draw[bluearr]
  (cf.south)
  -- ++(0,-.18)
  -| (mour.south);

\node[loopmain] at (5.84,.88)
  {self-supervised training};

\node[loopsub] at (5.84,.29)
  {no efficiency objective};

\end{tikzpicture}
}

\caption{\textbf{Teaching confidence instead of efficiency.}
Existing methods explicitly optimize for shorter reasoning traces; we supervise only confidence.}
\label{fig:concept}
\vspace{-0.5\baselineskip}
\end{wrapfigure}

Several lines of work have sought to improve the efficiency of reasoning in LLMs. Despite their differences, these approaches all make efficiency an explicit part of either training or inference. At inference time, early-stopping methods monitor signals such as confidence~\citep{deer,codestop}, uncertainty~\citep{eat}, or answer stability~\citep{answerconv} and terminate reasoning when further computation appears unnecessary. At training time, other approaches directly encourage shorter reasoning, for example through reinforcement learning with length penalties~\citep{dler,arora2025traininglanguagemodelsreason} or fine-tuning on selected concise reasoning trajectories~\citep{onpolicysft,munkhbat2025selftrainingelicitsconcisereasoning}. These approaches either directly control when reasoning stops or explicitly teach the model to produce shorter reasoning.

In this work, we study a different question: \emph{Can efficient reasoning emerge without directly training the model to produce shorter reasoning traces?} We introduce \textbf{ConfSFT}, a self-supervised fine-tuning procedure that trains a reasoning model to predict its confidence at intermediate points along its own reasoning trajectories. The confidence targets are derived entirely from the model's own token probabilities and require neither gold answers nor external judges. Crucially, confidence is the only supervised signal: reasoning tokens are masked from the loss, and the objective contains no term for reasoning length, efficiency, or stopping. At inference, the fine-tuned model uses standard generation, with no confidence elicitation or stopping mechanism.  Figure~\ref{fig:concept} illustrates this.

The motivation for this supervision comes from a simple observation: confidence provides a meaningful signal about the model's intermediate reasoning state. Across intermediate states, higher confidence is associated with more reliable and stable answers, while the expected benefit of further reasoning decreases. This makes confidence a natural signal to learn: it reflects both the reliability of the current answer and the potential value of further reasoning.

Learning this signal alone substantially shortens reasoning: ConfSFT reduces generated tokens by up to 25\% at matched accuracy across four model families and transferring from mathematical training problems to scientific and coding tasks. Its efficiency gains are comparable to training methods that explicitly optimize for shorter reasoning.

Our analyses further show that confidence prediction improves over training while generated tokens decrease and accuracy remains stable. A decomposition of reasoning traces into high-level reasoning behaviors shows that ConfSFT largely preserves the base models' reasoning composition, rather than achieving efficiency by selectively suppressing a particular type of reasoning. Controlled ablations further show that the supervision signal matters: alternative targets, including randomly shuffled confidence labels, produce weaker or no efficiency gains. Together, these results provide evidence that learning confidence itself can reshape reasoning toward greater efficiency.
\section{Related Work}

\textbf{Training LLMs for efficient reasoning.} A growing line of work directly trains reasoning models to use less computation. Several approaches use reinforcement learning objectives that explicitly favor shorter correct reasoning traces~\citep{dler,arora2025traininglanguagemodelsreason,yi2025shorterbetterguidingreasoningmodels}. Others fine-tune on self-generated concise trajectories, either by selecting short reasoning paths or filtering on-policy generations for correctness and conciseness~\citep{munkhbat2025selftrainingelicitsconcisereasoning,onpolicysft}. Other methods explicitly train models to adapt or terminate their reasoning through truncated rollouts, selective reasoning, or confidence-guided post-training~\citep{stepgrpo,sageRL,cat,jet,concise}. While these methods differ in how efficiency is encouraged, they all make shorter or more efficient reasoning an explicit part of the training procedure. In contrast, our approach supervises only confidence, without any explicit training signal that prefers shorter reasoning.

\textbf{Inference-time adaptive reasoning and early stopping.} A complementary line of work reduces reasoning cost by adapting computation at inference time. Several methods use confidence or uncertainty signals to determine when further reasoning is unnecessary~\citep{deer,codestop,eat,kim-etal-2026-think,thinkornot,fu2025deepthinkconfidence}. Others rely on the convergence or stability of intermediate answers to terminate reasoning~\citep{answerconv,mao2026earlystoppingchainofthoughtslarge}, while \citet{zhang2025reasoningmodelsknowtheyre} train probes on internal representations to detect intermediate answer correctness and enable early exit. Although these approaches use different signals, they all explicitly use an inference-time criterion to control how much reasoning is performed. In contrast, ConfSFT does not use confidence or any other stopping criterion during inference.

\textbf{Confidence and metacognitive learning.} A growing body of work incorporates confidence and metacognitive signals into training. \citet{verbalized_confidence} show that fine-tuning on verbalized confidence labels alone can induce changes in reasoning behavior, including emergent self-verification. Other work explicitly trains models to improve their confidence calibration, for example by optimizing verbalized confidence calibration through reinforcement learning~\citep{baniharouni2026rewardingdoubtreinforcementlearning,damani2026binaryrewardstraininglms}, shaping confidence dynamics to avoid premature commitment~\citep{prematureconfidence}, or using metacognitive self-evaluation as feedback during reinforcement learning~\citep{liu2026reinforcementlearningmetacognitivefeedback}. Confidence has also been incorporated into reward and alignment objectives to improve reasoning quality and reliability~\citep{rewardmodelingbeyondcorrectnes,caspo}. 
ConfSFT differs in both the source and role of the confidence signal. We derive confidence targets directly from the model's own token probabilities at intermediate reasoning states and train only to predict these targets, without gold correctness labels or external feedback. Rather than optimizing efficiency, correctness, or a desired confidence trajectory, we study whether this self-supervised confidence learning alone can induce more efficient reasoning.
\section{Confidence and the Value of Further Reasoning}
\label{sec:motivation}

\begin{figure}[t]
  \centering
  \includegraphics[width=\linewidth]{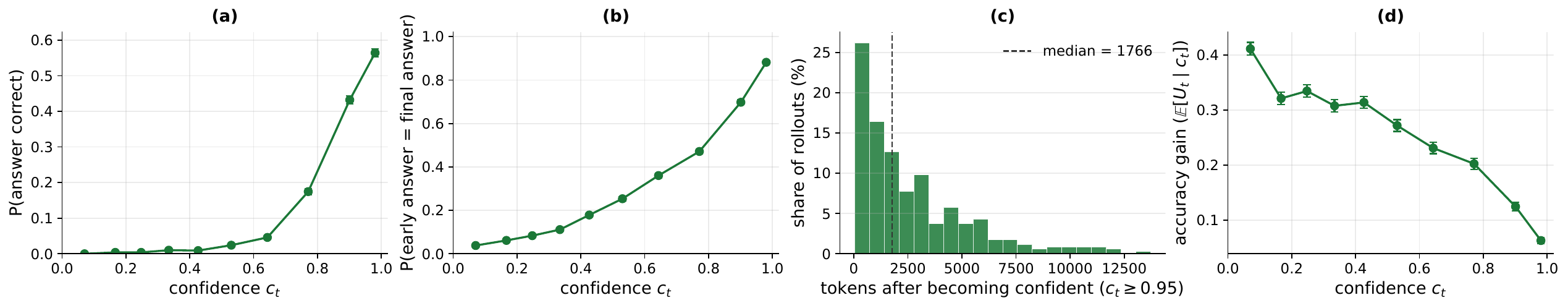}
  \caption{\textbf{(a)} Trial-answer correctness and \textbf{(b)} agreement with the final answer increase with confidence.
\textbf{(c)} The model often continues generating after first reaching $c_t \ge 0.95$.
\textbf{(d)} The expected accuracy gain from further reasoning decreases with confidence.
  }
  \label{fig:conf-waste-four}
\end{figure}

\textbf{What do we mean by confidence?} At an intermediate point in reasoning, we probe the model for its current answer. Given the reasoning generated so far, we append a fixed answer-elicitation prompt (e.g., \texttt{The final answer is \textbackslash boxed\{}) and greedily generate the model's current trial answer, as illustrated in Figure~\ref{fig:method}a (Step 2). Let $t$ index such an intermediate state and $a_t$ denote its trial answer. We define the corresponding \emph{confidence} $c_t$ as the geometric mean of the token probabilities assigned to $a_t$, i.e., its length-normalized likelihood. Importantly, this score is computed entirely from the model's own distribution and does not require knowing whether the trial answer is correct. The exact answer-elicitation prompts are provided in Appendix~\ref{app:prompts:probe}. 


We do not interpret this score as a calibrated probability of correctness: a confidence of $0.8$, for example, does not necessarily imply that the answer is correct $80\%$ of the time. Instead, we ask whether it provides a meaningful signal about the model's intermediate reasoning state and the value of continuing to reason.


\textbf{Confidence is informative about the reasoning state.} We evaluate confidence at intermediate reasoning states immediately preceding each \texttt{Wait} in 480 Nemotron traces on AIME~2024 (16 samples per problem) and group the resulting intermediate states into equal-sized confidence quantiles, with roughly 1,700 states per bin. 

As shown in Figure~\ref{fig:conf-waste-four}, higher confidence is strongly associated with a higher probability that the trial answer is correct. The relationship is even stronger with answer stability: as confidence increases, the trial answer becomes increasingly likely to already match the model's eventual final answer. We observe the same trends for Gemma in Appendix~\ref{app:cmaegemma}.

\textbf{Confidence tracks the value of further reasoning.} Models nevertheless often continue reasoning after reaching a high-confidence state. After first reaching $c_t \ge 0.95$, Nemotron generates a median of 1,766 additional tokens, with many trajectories continuing for more than 10K tokens (Figure~\ref{fig:conf-waste-four}).

To quantify whether this additional computation is useful, we define the accuracy gain from continuing at an intermediate state $t$ as
\begin{equation}
U_t
=
\mathbbm{1}[a_{\mathrm{final}}\ \text{is correct}]
-
\mathbbm{1}[a_t\ \text{is correct}],
\end{equation}
where $a_t$ is the trial answer at $t$ and $a_{\mathrm{final}}$ is the answer after completing the reasoning trajectory. Thus, $U_t>0$ when further reasoning corrects the current answer and $U_t<0$ when it changes a correct intermediate answer into an incorrect final answer. As shown in Figure~\ref{fig:conf-waste-four}, the expected gain from continued reasoning decreases sharply with confidence: further reasoning is most beneficial at low-confidence states and provides little average improvement once confidence is high.

Together, these observations suggest that confidence provides a self-supervised signal about both the model's current reasoning state and the value of additional computation. Prior work has used the same confidence signals to explicitly control reasoning, either through inference-time early stopping~\citep{deer,codestop} or by truncating rollouts during training~\citep{stepgrpo}. We instead ask a different question: what happens if the model is simply trained to predict this signal at intermediate reasoning states? In the following sections, we show that confidence supervision alone can substantially reduce generated tokens, without truncating trajectories or explicitly encouraging shorter reasoning.

\section{Method}
\label{sec:method}

\begin{figure}[t]
  \centering
\tikzset{
  arr/.style={-{Latex[length=1.5mm]}, semithick, draw=textgray},
  route/.style={-{Latex[length=1.5mm]}, semithick, draw=textgray, rounded corners=3pt},
  stepbox/.style={
    draw=bordergray, fill=white, rounded corners=3pt,
    inner xsep=2pt, inner ysep=2pt, align=center,
    font=\fontsize{6.6}{7.4}\selectfont\bfseries, text=black,
    minimum height=0.56cm, minimum width=2.10cm,
  },
  stepnum/.style={
    circle, fill=stepink, text=white, font=\fontsize{5.6}{5.6}\selectfont\bfseries,
    inner sep=0pt, minimum size=0.29cm,
  },
  chip/.style={
    draw=bordergray, rounded corners=1.8pt,
    inner xsep=2.4pt, inner ysep=1.6pt,
    font=\ttfamily\fontsize{7}{8}\selectfont,
  },
  tok/.style={chip, fill=answergreen, font=\ttfamily\fontsize{7}{8}\selectfont\bfseries, minimum width=0.30cm},
  lmbox/.style={
    draw=bordergray, fill=lmblue, rounded corners=3pt,
    font=\fontsize{7}{8}\selectfont\bfseries, minimum height=0.50cm, minimum width=2.40cm,
  },
  card/.style={draw=bordergray, rounded corners=3pt, fill=white, inner sep=5pt, align=left},
  panel/.style={draw=panelrule, rounded corners=3pt, fill=black!2, inner sep=4pt},
  ttl/.style={
    rounded corners=2pt, fill=stepink, text=white, anchor=west,
    font=\fontsize{7.4}{8.6}\selectfont\bfseries, inner xsep=4pt, inner ysep=2.2pt,
  },
  tag/.style={font=\fontsize{6}{7}\selectfont, text=textgray},
  trace/.style={font=\fontsize{6}{7.2}\selectfont, align=left},
}
\newcommand{\wt}{\colorbox{decisionlilac}{\strut\,\textbf{Wait}\,}}
\newcommand{\ansbox}{\colorbox{answergreen}{\strut\,\textbf{55}\,}}

\begin{tikzpicture}[x=1cm, y=1cm]

\node[ttl] (patitle) at (0.18,5.16) {(a) Training};

\node[stepbox, minimum width=3.95cm] (s1) at (2.28,4.60) {Generate\\[-2pt]rollouts};
\node[stepbox, minimum width=3.95cm] (s2) at (7.03,4.60) {Construct\\[-2pt]confidence labels};
\node[stepbox, minimum width=3.95cm] (s3) at (11.78,4.60) {Confidence-supervised\\[-2pt]fine-tuning};
\foreach \i/\n in {s1/1, s2/2, s3/3}
  \node[stepnum] at (\i.north west) {\n};
\draw[arr] (s1) -- (s2);
\draw[arr] (s2) -- (s3);
\draw[route] (s3.south) -- ++(0,-0.24) -| (s1.south);
\node[tag, fill=black!2, inner xsep=3pt] at (7.03,4.06) {repeat};

\node[stepnum] (n1) at (0.48,3.56) {1};
\node[chip, fill=promptgreen, anchor=west] (pq) at (0.90,3.56) {Problem};
\node[lmbox, anchor=west, right=8pt of pq, minimum width=1.90cm, minimum height=0.42cm,
      font=\fontsize{6.4}{7.4}\selectfont\bfseries] (lm0) {Reasoning model};
\draw[arr] (pq.east) -- (lm0.west);
\node[chip, fill=reasonblue, anchor=west, right=8pt of lm0] (r1) {$\textstyle\sum 2^{a-1}=2024$};
\draw[arr] (lm0.east) -- (r1.west);
\node[chip, fill=decisionlilac, font=\ttfamily\fontsize{7}{8}\selectfont\bfseries, anchor=west, right=3pt of r1] (w1) {Wait};
\node[chip, fill=reasonblue, anchor=west, right=3pt of w1] (r2) {$2024=11111101000_2$};
\node[chip, fill=decisionlilac, font=\ttfamily\fontsize{7}{8}\selectfont\bfseries, anchor=west, right=3pt of r2] (w2) {Wait};
\node[chip, fill=reasonblue, anchor=west, right=3pt of w2] (r3) {sum $=55$};

\node[stepnum] at (0.48,2.66) {2};
\node[chip, fill=reasonblue, anchor=west] (ctx3) at (0.90,2.66) {$\textstyle\sum 2^{a-1}=2024$};
\node[chip, fill=promptgreen, font=\ttfamily\fontsize{6.4}{7.6}\selectfont, anchor=west, right=3pt of ctx3]
  (cue) {Final answer is:};
\node[tag, below=1.2pt of ctx3] (t31) {reasoning so far};
\node[tag, below=1.2pt of cue] (t32) {answer-elicitation prompt};

\draw[route] (w1.south) -- ++(0,-0.26) -| ($(ctx3.north)+(0.30,0)$);

\node[lmbox, anchor=west, right=8pt of cue] (lm) {Reasoning model};
\draw[arr] (cue.east) -- (lm.west);

\node[tok, anchor=west, right=10pt of lm] (tA) {5};
\node[tok, anchor=west, right=17pt of tA] (tB) {5};
\draw[arr] (lm.east) -- (tA.west);
\draw[arr] (tA.east) -- (tB.west);
\node[tag] (argen) at ($(tA.north)!0.5!(tB.north)+(-0.10,0.23)$) {one token at a time};
\node[tag, below=1.2pt of tA] (t33) {$p_1{=}.80$};
\node[tag, below=1.2pt of tB] (t34) {$p_2{=}.65$};

\draw[arr] ($(tB.east)+(0.36,0)$) -- ++(0.95,0);
\node[tag] at ($(tB.east)+(0.84,0.20)$) {geometric mean};
\node[chip, fill=targetorange, font=\ttfamily\fontsize{7}{8}\selectfont\bfseries, anchor=west]
  (conf) at ($(tB.east)+(1.58,0)$) {72\%};

\node[stepnum] (n4) at (0.48,1.66) {3};
\node[chip, fill=promptgreen, anchor=west] (mq) at (0.90,1.66) {Problem};
\node[chip, fill=reasonblue, anchor=west, right=3pt of mq] (mr) {$\textstyle\sum 2^{a-1}=2024$};
\node[chip, fill=maskgray, align=left, font=\ttfamily\fontsize{5.8}{6.9}\selectfont,
      anchor=west, right=3pt of mr] (pref)
  {From 0\% (very low) to 100\% (very high),\\my confidence in the answer so far is};
\node[chip, fill=targetorange, font=\ttfamily\fontsize{7}{8}\selectfont\bfseries, anchor=west, right=4pt of pref]
  (labv) {72\%};

\draw[decorate, decoration={brace, amplitude=2.6pt, mirror}, draw=bordergray]
  ($(mq.south west)+(0,-0.09)$) -- ($(pref.south east)+(0,-0.09)$)
  node[midway, below=3pt, tag] (mask) {masked};
\node[tag, text=losstext] (lossl) at ($(labv.south)+(0,-0.26)$) {\bfseries loss};

\draw[route, losstext] (conf.south) -- ++(0,-0.48) -| (labv.north);

\coordinate (aL) at (0.10,4.60);
\coordinate (aR) at (13.87,4.60);
\begin{scope}[on background layer]
\node[panel, fit=(patitle)(s1)(s3)(n1)(r3)(conf)(mask)(lossl)(aL)(aR)] {};
\end{scope}

\node[ttl] (pbtitle) at (0.18,0.06) {(b) Inference};

\node[card, fill=promptgreen, anchor=north west, text width=13.22cm, font=\fontsize{6}{7.2}\selectfont]
  (prob) at (0.30,-0.28) {\textbf{Problem.} Exactly $2024$ finite nonempty sets $B$ of positive integers
  have their maximum in $A$. Find $\textstyle\sum_{a\in A} a$.};

\node[card, anchor=north west, trace] (base) at (0.30,-0.78) {%
  \begin{minipage}[t][2.52cm][t]{6.28cm}
  \setlength{\fboxsep}{2pt}%
  \textbf{\fontsize{7.6}{9}\selectfont Base}\hfill \textbf{\fontsize{7.6}{9}\selectfont $\approx$14K tokens}\\[2pt]
  \tikz[baseline]{%
    \fill[reasonblue, draw=bordergray, rounded corners=1pt] (0,0) rectangle (1.00,0.14);
    \fill[wastepink, draw=bordergray, rounded corners=1pt] (1.00,0) rectangle (6.28,0.14);}\\[3pt]
  Each $a$ gives $2^{a-1}$ sets, so $\sum 2^{a-1}=2024$.\\[2pt]
  \colorbox{wastepink}{\parbox{5.98cm}{%
    \wt{} write $2024$ in binary: $11111101000_2$.\\[-1pt]
    \wt{} recheck $1024{+}512{+}\dots{+}8=2024$.\\[-1pt]
    \textit{\dots the same fact re-derived six times \dots}}}\\[3pt]
  Final answer: \ansbox
  \end{minipage}};

\node[card, anchor=north west, trace] (ours) at (7.24,-0.78) {%
  \begin{minipage}[t][2.52cm][t]{6.28cm}
  \setlength{\fboxsep}{2pt}%
  \textbf{\fontsize{7.6}{9}\selectfont Ours}\hfill \textbf{\fontsize{7.6}{9}\selectfont $\approx$3K tokens}\\[2pt]
  \tikz[baseline]{%
    \fill[reasonblue, draw=bordergray, rounded corners=1pt] (0,0) rectangle (1.00,0.14);
    \fill[answergreen, draw=bordergray, rounded corners=1pt] (1.00,0) rectangle (1.35,0.14);}\\[3pt]
  Each $a$ gives $2^{a-1}$ sets, so $\sum 2^{a-1}=2024$.\\
  $2024=11111101000_2\Rightarrow$ exponents $\{10,9,8,7,6,5,3\}$.\\
  So $A=\{11,10,9,8,7,6,4\}$.\\[2pt]
  \colorbox{answergreen}{\parbox{5.98cm}{%
    \wt{} check $11{+}10{+}9{+}8{+}7{+}6{+}4=55$.}}\\[3pt]
  Final answer: \ansbox
  \end{minipage}};

\coordinate (bL) at (0.10,-0.30);
\coordinate (bR) at (13.87,-0.30);
\begin{scope}[on background layer]
\node[panel, fit=(pbtitle)(prob)(base)(ours)(bL)(bR)] {};
\end{scope}

\end{tikzpicture}
  \vspace{-17pt}
  \caption{\textbf{Overview of ConfSFT.}
\textbf{(a)} The policy generates reasoning rollouts, constructs self-supervised confidence labels at intermediate states, and is fine-tuned only on the confidence targets. \textbf{(b)} At inference, the fine-tuned policy uses standard generation but can produce substantially shorter reasoning while reaching the same correct answer.}
  \vspace{-20pt}
  \label{fig:method}
\end{figure}
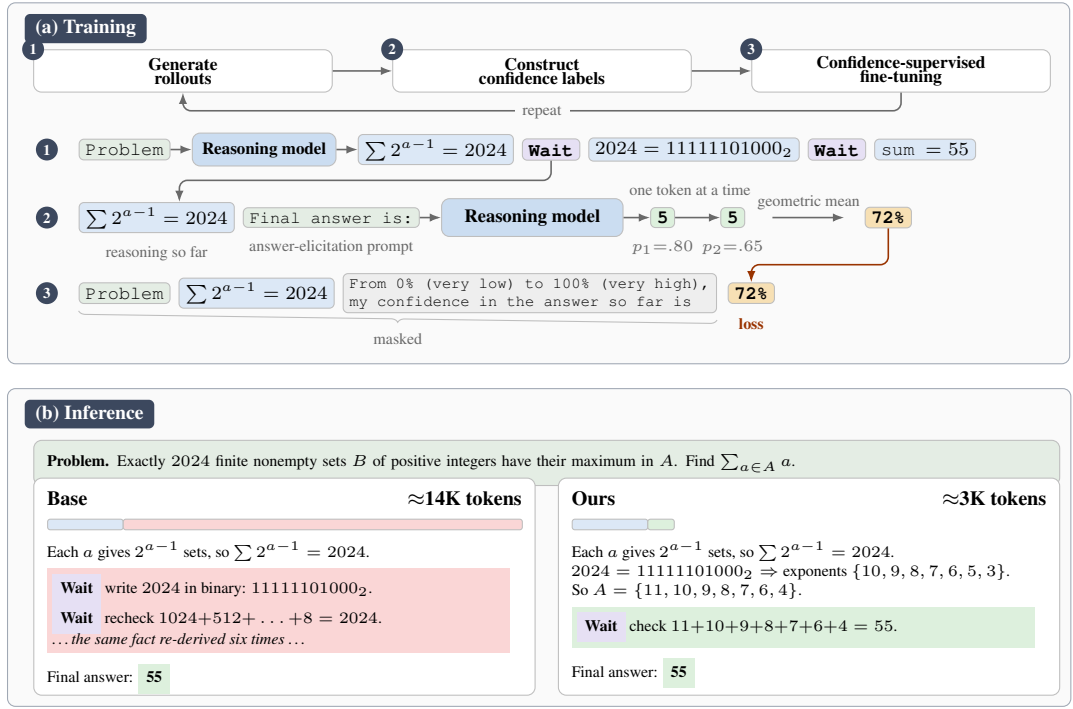

Motivated by the observations in Section~\ref{sec:motivation}, we introduce \textbf{ConfSFT}, a self-supervised procedure for fine-tuning reasoning models with confidence supervision at intermediate reasoning states. As illustrated in Figure~\ref{fig:method}, ConfSFT repeatedly generates reasoning trajectories from the current policy, constructs confidence labels from intermediate states, and fine-tunes the model to predict these labels. We describe each step below.


\paragraph{Generating reasoning rollouts.} We begin with a small subset $\mathcal{D}_{\mathrm{train}}$ of the training problems. For each problem $x \in \mathcal{D}_{\mathrm{train}}$, we use the current policy $\pi_\theta$ to generate one or more reasoning trajectories $\tau=(y_1,\ldots,y_T)$. These are standard reasoning rollouts, generated without confidence elicitation or any instruction to reason efficiently. The prompts are provided in Appendix~\ref{app:prompts:reasoning}. 

\paragraph{Constructing confidence labels.} For each rollout, we identify a set of intermediate \emph{decision points} using a fixed textual marker. By default, we use occurrences of \texttt{Wait} and take the reasoning prefix immediately preceding each occurrence as an intermediate reasoning state. Let $t$ index such a decision point and $y_{<t}$ denote the corresponding reasoning prefix. When a trajectory contains many decision points, we retain at most $M$, selected approximately uniformly across the trajectory, to prevent a single rollout from contributing disproportionately to the training set.

To construct a confidence target for $y_{<t}$, we append a fixed answer-elicitation prompt $q$ and greedily generate a trial answer $a=(a_1,\ldots,a_n)$. We define its confidence as the geometric mean of the token probabilities assigned to the generated answer:
\begin{equation}
c_t =
\exp\left(
\frac{1}{n}\sum_{i=1}^{n}
\log \pi_\theta
\left(a_i \mid x, y_{<t}, q, a_{<i}\right)
\right).
\label{eq:confidence}
\end{equation}
This corresponds to the length-normalized likelihood of the trial answer. Importantly, $c_t$ is computed entirely from the model's own distribution and does not require the gold answer.


We represent confidence as a textual percentage so that it can be learned using the model's standard next-token prediction objective. Specifically, we quantize $c_t\in[0,1]$ onto a grid of $K=50$ percentage levels, corresponding to $2\%$ increments, and round each score up to the smallest grid point greater than or equal to it. The resulting textual percentage $\ell_t$ serves as the supervision target for decision point $t$.

\paragraph{Confidence-supervised fine-tuning.}

For each decision point, we construct one training example by concatenating the problem prompt, the corresponding reasoning prefix $y_{<t}$, a fixed natural-language \emph{priming prefix} $p$, and the quantized confidence label $\ell_t$:
\begin{equation}
    s
    =
    \underbrace{
        \mathrm{prompt}(x) \oplus y_{<t} \oplus p
    }_{\text{context, masked}}
    \oplus
    \underbrace{
        \ell_t
    }_{\text{supervised}}
    \label{eq:sequence}
\end{equation}
where $\oplus$ denotes concatenation. We use the fixed priming prefix:
\begin{center}
\scalebox{0.85}{%
\ttfamily From 0\% (very low) to 100\% (very high), my confidence in the answer so far is%
}
\end{center}

Let $S(s)$ denote the token positions corresponding to the confidence label $\ell_t$. We optimize the standard next-token cross-entropy only over these positions:
\begin{equation}
    \mathcal{L}(\theta)
    =
    -\mathbb{E}_{(x,y,t)}
    \left[
        \sum_{j \in S(s)}
        \log \pi_\theta
        \left(
            s_j \mid s_{<j}
        \right)
    \right].
    \label{eq:loss}
\end{equation}

All other tokens, including the problem prompt, reasoning prefix, and priming prefix, are masked from the loss. Thus, the model is trained only to predict the confidence label conditioned on an intermediate reasoning state; it is not trained to reproduce the sampled reasoning trajectory or to generate confidence text within its chain of thought. The objective contains no term for reasoning length, stopping, or efficiency.

\paragraph{Iterative on-policy training.}
Because both reasoning traces and confidence targets are policy-generated, we refresh them as the policy changes. At round $r$, we sample rollouts from a fresh subset of training problems using $\pi_{\theta_r}$, construct confidence labels at decision points, and fine-tune on the resulting examples to obtain $\pi_{\theta_{r+1}}$ (Figure~\ref{fig:method}a). Later rounds generally yield more efficient policies. We evaluate the policy after each round on a fixed held-out validation set.

\paragraph{Inference.}
Inference is unchanged from the base model. The fine-tuned model uses the same decoding procedure, with no confidence instruction, answer-elicitation cue, priming prefix, verifier, or early-exit mechanism. The fine-tuned models also do not spontaneously emit confidence values or the priming prefix within their reasoning traces. Thus, any change in reasoning length arises from the fine-tuned policy itself rather than an inference-time intervention. Figure~\ref{fig:method}b illustrates this behavior on an AIME problem, where the fine-tuned policy reaches the same correct answer with substantially shorter reasoning.

\section{Experiments}
\subsection{Setup}
\textbf{Models.}
We evaluate ConfSFT on four reasoning model families of different scales: Gemma-4-E2B~\citep{gemmateam2026gemma4}, Qwen3-4B~\citep{qwen3technicalreport}, Nemotron-Nano-8B~\citep{nemotron}, and GPT-OSS-20B~\citep{gptoss}.

\textbf{Training Data.}
We train on problems from AIME2000--2023~\citep{aime19832024}. AIME2024 is held out for validation; no test benchmark is used during training or model selection.

\textbf{Evaluation.} We evaluate on AIME2025~\citep{aime2025}, GSM8K~\citep{gsm8k}, GPQA-Diamond~\citep{gpqa}, HumanEval~\citep{humaneval}, and LiveCodeBench~\citep{jain2024livecodebench}. We sample 16 completions per problem and report accuracy and average generated tokens across them. The full evaluation protocol is provided in Appendix~\ref{app:eval}.

\textbf{Training Details.}
We apply the ConfSFT procedure described in Section~\ref{sec:method}. We partition the training problems into eight groups, allowing up to eight training rounds, and sample eight rollouts per training problem and 16 per validation problem. Full training hyperparameters and implementation details are provided in Appendix~\ref{app}.

\textbf{Baselines.}
We compare against the corresponding base model and DEER~\citep{deer}, an inference-time early-stopping method that uses confidence to terminate reasoning. As training-time baselines, we evaluate A\&Z~\citep{arora2025traininglanguagemodelsreason}, which explicitly incorporates response length into the RL objective, on Nemotron and Gemma, and On-Policy SFT~\citep{onpolicysft} on Nemotron, Gemma, and Qwen. Full baseline training details and more experiments are provided in Appendix~\ref{app:baselines}.

\textbf{Metrics.} We report accuracy (\textbf{Acc.}), average number of generated tokens (\textbf{Avg.\ Tokens}), and token reduction (\textbf{Token Red.}), defined as the percentage reduction in average generated tokens relative to the corresponding base model. We also report pass@8 (\textbf{P@8}) in Appendix~\ref{app:eval}

\subsection{Main Results}
\label{sec:results}

\begin{table}[t]
  \centering
  \caption{\textbf{Main results.} ConfSFT reduces generated tokens while maintaining base-model accuracy across model families and task domains.}
  \label{tab:main}
  \vspace{2pt}
  \setlength{\tabcolsep}{2.2pt}
  \renewcommand{\arraystretch}{1.12}
  \arrayrulecolor{tablerule}
  \scriptsize
  \newcommand{\hacc}{\multicolumn{1}{c}{Acc$\uparrow$}}
  \newcommand{\htok}{\multicolumn{1}{c}{Avg.\ Tokens$\downarrow$}}
  \newcommand{\hsav}{\multicolumn{1}{c}{Token Red.}}
  \newcommand{\haccG}{\multicolumn{1}{!{\color{tablerule}\vrule}c}{Acc$\uparrow$}}
  \newcommand{\dom}[2]{\multicolumn{#1}{c!{\color{tablerule}\vrule}}{\textsc{#2}}}
  \newcommand{\domend}[2]{\multicolumn{#1}{c}{\textsc{#2}}}
  \newcommand{\bn}[1]{\multicolumn{2}{c!{\color{tablerule}\vrule}}{#1}}
  \newcommand{\bnend}[1]{\multicolumn{2}{c}{#1}}
  \newcommand{\ts}[2]{#1\,\textcolor{savedgreen}{\tiny(#2\%)}}
  \newcommand{\sv}[1]{\textcolor{savedgreen}{#1\%}}
  \resizebox{\linewidth}{!}{%
  \begin{tabular}{@{}l !{\color{tablerule}\vrule}
      c@{\hspace{4pt}}c !{\color{tablerule}\vrule}
      c@{\hspace{4pt}}c !{\color{tablerule}\vrule}
      c@{\hspace{4pt}}c !{\color{tablerule}\vrule}
      c@{\hspace{4pt}}c !{\color{tablerule}\vrule}
      c@{\hspace{4pt}}c !{\color{tablerule}\vrule}
      c@{\hspace{4pt}}c@{}}
    \toprule
    \multirow{3}{*}{Method}
    & \dom{4}{Math} & \dom{2}{Science} & \dom{4}{Coding} & \domend{2}{Average} \\
    & \bn{AIME2025} & \bn{GSM8K} & \bn{GPQA-Diamond} & \bn{LiveCodeBench} & \bn{HumanEval} & \bnend{} \\
    \cmidrule(lr){2-3}\cmidrule(lr){4-5}\cmidrule(lr){6-7}\cmidrule(lr){8-9}\cmidrule(lr){10-11}\cmidrule(lr){12-13}
    & \hacc & \htok & \haccG & \htok
           & \haccG & \htok & \haccG & \htok
           & \haccG & \htok & \haccG & \hsav \\
    \midrule
    \multicolumn{13}{@{}l}{\textit{Nemotron-Nano-8B}} \\
    Base  & 42.5 & 11,325 & 91.8 & 1,192 & 52.5 & 7,338 & 44.2 & 11,761 & 89.7 & 3,600 & 64.1 & \sv{0.0} \\
    DEER\citep{deer}  & 39.8 & \ts{9,076}{-19.9} & 89.2 & \ts{765}{-35.8} & 49.8 & \ts{5,712}{-22.2} & 17.4 & \ts{1,731}{-85.3} & 47.1 & \ts{511}{-85.8} & 48.7 & \sv{-49.8} \\
    On-Policy SFT\citep{onpolicysft} & 39.0 & \ts{10,487}{-7.4} & 91.1 & \ts{999}{-16.2} & 50.3 & \ts{6,146}{-16.2} & 41.8 & \ts{11,364}{-3.4} & 89.3 & \ts{3,174}{-11.8} & 62.3 & \sv{-11.0} \\
    A\&Z\citep{arora2025traininglanguagemodelsreason} & 41.5 & \ts{9,644}{-14.8} & 91.0 & \ts{1,218}{+2.2} & 50.9 & \ts{6,738}{-8.2} & 44.5 & \ts{11,346}{-3.5} & 88.4 & \ts{3,489}{-3.1} & 63.3 & \sv{-5.5} \\
    \rowcolor{oursrow}
    \textbf{ConfSFT (ours)} & 45.6 & \ts{10,153}{-10.4} & 91.6 & \ts{1,010}{-15.3} & 52.8 & \ts{6,278}{-14.4} & 45.1 & \ts{11,151}{-5.2} & 90.8 & \ts{3,225}{-10.4} & 65.2 & \sv{-11.1} \\
    \midrule
\multicolumn{13}{@{}l}{\textit{Gemma-4-E2B}} \\

Base
& 35.0 & 7,284
& 91.2 & 956
& 42.6 & 3,294
& 42.1 & 7,543
& 93.2 & 2,210
& 60.8 & \sv{0.0} \\

DEER\citep{deer}
& 24.2 & \ts{4,924}{-32.4}
& 90.4 & \ts{912}{-4.6}
& 36.5 & \ts{2,432}{-26.2}
& 31.3 & \ts{5,333}{-29.3}
& 82.9 & \ts{1,905}{-13.8}
& 53.1 & \sv{-21.3} \\

On-Policy SFT\citep{onpolicysft}
& 27.3 & \ts{5,601}{-23.1}
& 90.9 & \ts{1,049}{+9.7}
& 42.1 & \ts{3,587}{+8.9}
& 41.0 & \ts{7,711}{+2.2}
& 93.2 & \ts{2,318}{+4.9}
& 58.9 & \sv{+0.5} \\

A\&Z\citep{arora2025traininglanguagemodelsreason}
& 35.4 & \ts{7,018}{-3.7}
& 91.3 & \ts{963}{+0.7}
& 41.5 & \ts{3,224}{-2.1}
& 41.2 & \ts{7,291}{-3.3}
& 93.6 & \ts{2,200}{-0.5}
& 60.6 & \sv{-1.8} \\

\rowcolor{oursrow}
\textbf{ConfSFT (ours)}
& 32.1 & \ts{5,987}{-17.8}
& 91.4 & \ts{938}{-1.9}
& 40.8 & \ts{2,968}{-9.9}
& 42.4 & \ts{6,403}{-15.1}
& 93.8 & \ts{2,059}{-6.8}
& 60.1 & \sv{-10.3} \\
    \midrule
    \multicolumn{13}{@{}l}{\textit{Qwen3-4B}} \\
    Base  & 59.2 & 13,268 & 94.8 & 2,292 & 53.0 & 9,075 & 46.8 & 14,954 & 93.3 & 3,495 & 69.4 & \sv{0.0} \\
    DEER\citep{deer}  & 40.4 & \ts{9,086}{-31.5} & 84.8 & \ts{580}{-74.7} & 47.1 & \ts{3,621}{-60.1} & 25.8 & \ts{10,374}{-30.6} & 50.8 & \ts{756}{-78.4} & 49.8 & \sv{-55.0} \\
    On-Policy SFT\citep{onpolicysft} & 58.8 & \ts{11,405}{-14.0} & 95.1 & \ts{1,414}{-38.3} & 53.9 & \ts{7,694}{-15.2} & 45.9 & \ts{14,518}{-2.9} & 93.1 & \ts{2,944}{-15.7} & 69.3 & \sv{-17.2} \\
    \rowcolor{oursrow}
    \textbf{ConfSFT (ours)} & 58.8 & \ts{11,365}{-14.3} & 94.5 & \ts{1,710}{-25.4} & 55.8 & \ts{6,838}{-24.6} & 45.0 & \ts{12,122}{-18.9} & 93.3 & \ts{3,054}{-12.6} & 69.5 & \sv{-19.2} \\
    \midrule
    \multicolumn{13}{@{}l}{\textit{gpt-oss-20b}} \\
    Base  & 52.7 & 5,802 & 94.0 & 482 & 61.0 & 3,732 & 53.5 & 5,034 & 95.9 & 917 & 71.4 & \sv{0.0} \\
    DEER\citep{deer}  & 52.7 & \ts{5,807}{+0.1} & 94.0 & \ts{482}{+0.1} & 61.0 & \ts{3,753}{+0.6} & 51.1 & \ts{4,824}{-4.2} & 90.6 & \ts{832}{-9.2} & 69.9 & \sv{-2.5} \\
    \rowcolor{oursrow}
    \textbf{ConfSFT (ours)} & 53.3 & \ts{5,235}{-9.8} & 94.1 & \ts{437}{-9.2} & 61.0 & \ts{3,112}{-16.6} & 50.7 & \ts{4,393}{-12.7} & 95.0 & \ts{850}{-7.2} & 70.8 & \sv{-11.1} \\
    \bottomrule
  \end{tabular}%
  }
\end{table}

Table~\ref{tab:main} shows that ConfSFT consistently reduces generated tokens while maintaining base-model accuracy across all four model families. ConfSFT reduces tokens by 11.1\%, 10.3\%, 19.2\%, and 11.1\% on Nemotron, Gemma, Qwen, and GPT-OSS, respectively. In Appendix~\ref{app:ci}, we report confidence intervals and show that the token reductions are statistically significant, while the changes in accuracy are not. Importantly, although ConfSFT is trained only on AIME problems, the efficiency gains transfer beyond mathematics to scientific reasoning and coding benchmarks.

\textbf{Comparison with explicit efficiency training.} We compare ConfSFT with two training-time approaches: A\&Z~\citep{arora2025traininglanguagemodelsreason} and On-Policy SFT~\citep{onpolicysft}. Across the model families where these baselines are evaluated, ConfSFT achieves comparable or larger efficiency gains despite receiving no supervision that favors shorter reasoning. For example, on Qwen, On-Policy SFT reduces tokens by 17.2\% at 69.3 average accuracy, compared with 19.2\% at 69.5 accuracy for ConfSFT. On Nemotron, ConfSFT similarly achieves substantial token reduction while maintaining base-model accuracy. These results show that confidence-only supervision can yield efficiency gains comparable to methods explicitly designed to shorten reasoning.

\textbf{Comparison with inference-time early stopping.} We additionally compare against DEER~\citep{deer}, which uses confidence to explicitly terminate reasoning at inference time. DEER can produce larger token reductions, but its effectiveness varies considerably across tasks and can come with large accuracy losses. This is particularly pronounced on coding benchmarks: with Nemotron, LiveCodeBench accuracy drops from 44.2\% to 17.4\% and HumanEval accuracy from 89.7\% to 47.1\%. In contrast, ConfSFT reduces generated tokens without applying an inference-time stopping rule and preserves accuracy across domains. Additional baseline details are provided in Appendix~\ref{app:baselines}.


\subsection{Confidence Learning and Efficiency Dynamics}

We next examine whether the model actually learns the confidence-prediction task, and how this relates to the efficiency gains during training. We measure prediction quality using \emph{confidence mean absolute error} (C-MAE), the mean absolute difference between the model's predicted confidence and the quantized self-supervised target on held-out reasoning states. Predictions are elicited using the same priming prefix as in Section~\ref{sec:method}; lower C-MAE indicates better prediction of the confidence targets.

\begin{figure}[t]
  \centering
  \includegraphics[width=\linewidth]{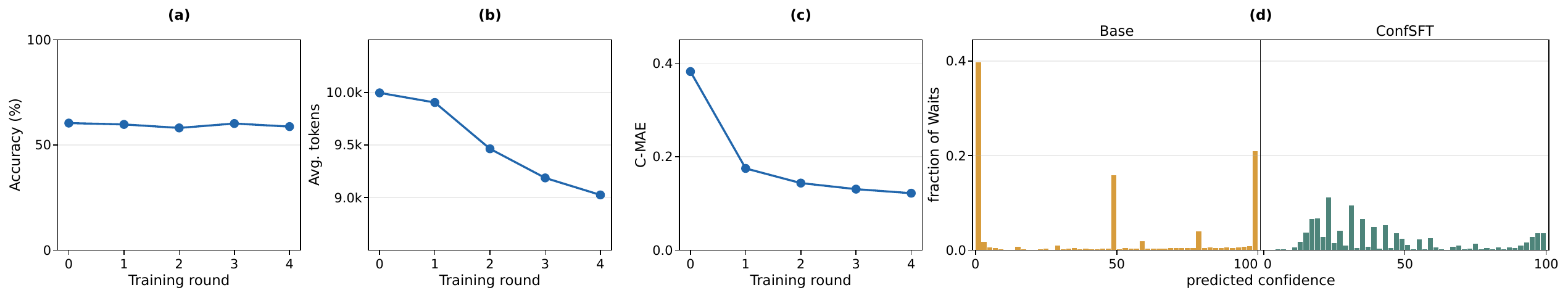}
  \caption{\textbf{Training Dynamics.} \textbf{(a)} Accuracy remain stable. \textbf{(b)} Generated tokens decrease. \textbf{(c)} Confidence predictions improved. \textbf{(d)} Confidence predictions become more graded.}
  \label{fig:cmae-over-epochs}
\end{figure}

Figure~\ref{fig:cmae-over-epochs}(a--c) shows the training dynamics for Nemotron on the validation set. As training progresses, C-MAE decreases substantially and average generated tokens fall, while accuracy remains approximately unchanged. Thus, the model learns to predict the confidence targets at the same time that its reasoning becomes more efficient.

Figure~\ref{fig:cmae-over-epochs}(d) further shows how the predicted confidence distribution changes. Before training, the model concentrates its predictions on only a few confidence levels. After ConfSFT, predictions span a much broader range, suggesting that the model learns to make more graded confidence judgments. We observe similar confidence-learning dynamics for Gemma in Appendix~\ref{app:gemma-analysis}.

\begin{figure}[t]
  \centering
  \includegraphics[width=\linewidth]{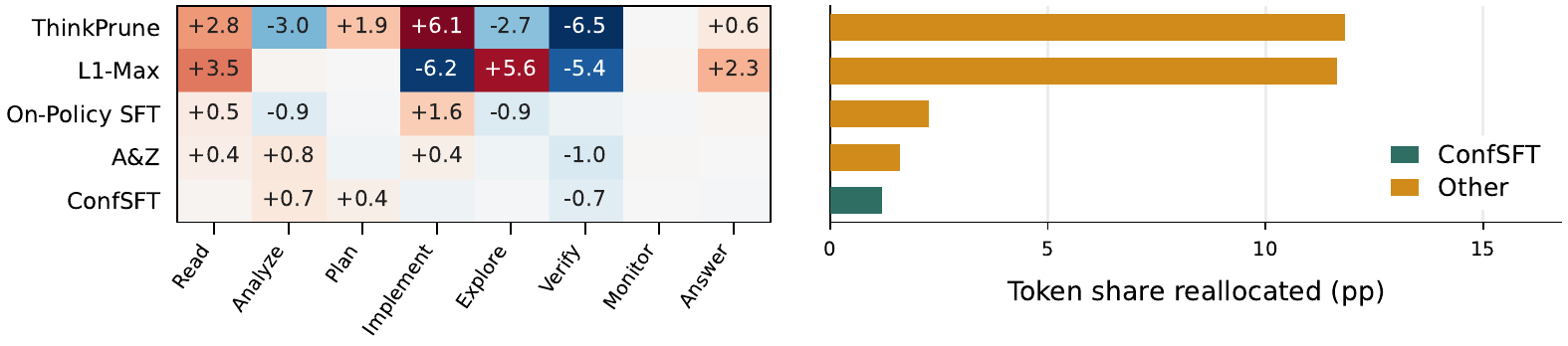}
  \vspace{-22pt}
  \caption{\textbf{Different efficiency methods reshape reasoning in different ways.} Following the Schoenfeld-style episode taxonomy, we compare each efficient policy with its corresponding base policy. \textbf{Left:} change in reasoning-token share for each episode, in percentage points. \textbf{Right:} total variation distance between the before- and after-training episode distributions.}
  \label{fig:schoenfeld}
\end{figure}



\subsection{Reasoning Composition under Efficiency Training}

Prior work has used Schoenfeld's Episode Theory to decompose model reasoning into different cognitive steps~\citep{Schoenfeld_emnlp,Schoenfeld}. In particular, ThinkARM~\citep{Schoenfeld} annotates reasoning at the sentence level using eight categories---\emph{Read}, \emph{Analyze}, \emph{Plan}, \emph{Implement}, \emph{Explore}, \emph{Verify}, \emph{Monitor}, and \emph{Answer}---and shows that efficiency methods can alter reasoning in qualitatively different ways, selectively changing the amount of computation allocated to different episodes rather than simply shortening all reasoning uniformly.

Following this framework, we annotate reasoning traces before and after efficiency training and compare the fraction of reasoning tokens assigned to each episode. We compare ConfSFT with several explicit efficiency-training methods, including L1-Max~\citep{l1max}, ThinkPrune~\citep{thinkprune}, A\&Z~\citep{arora2025traininglanguagemodelsreason}, and On-Policy SFT~\citep{onpolicysft}. We perform this analysis on AIME2025 with 16 sampled completions per problem for each method, using publicly available checkpoints when available and our reproductions otherwise. Full annotation, checkpoint, and details are provided in Appendix~\ref{app:schoenfeld}.

Figure~\ref{fig:schoenfeld} shows the change in episode composition relative to each method's corresponding base model. The heatmap reports the signed change in token share for each episode. We summarize the overall change as
\begin{equation}
D_{\mathrm{TV}}
=
\frac{1}{2}\sum_{k=1}^{8}
\left|
p_k^{\mathrm{after}}-p_k^{\mathrm{base}}
\right|,
\end{equation}
where $p_k$ is the fraction of reasoning tokens assigned to episode $k$. Equivalently, $D_{\mathrm{TV}}$ measures how much reasoning-token share is reallocated across episode categories after training.

Consistent with prior work~\citep{Schoenfeld}, different efficiency methods produce markedly different changes in reasoning composition. L1-Max, ThinkPrune, and On-Policy SFT substantially redistribute computation across episodes, with large shifts in categories such as \emph{Implement}, \emph{Explore}, and \emph{Verify}. In contrast, ConfSFT induces relatively small changes in episode shares, largely preserving the reasoning composition of the corresponding base policies. This suggests that ConfSFT does not achieve efficiency by selectively suppressing a particular reasoning behavior, but rather reduces computation while preserving the relative allocation across high-level reasoning episodes.

\section{Ablations}
\label{sec:ablations}

\subsection{What supervision signal matters?}

We test whether the efficiency gains of ConfSFT arise specifically from learning confidence, or more generally from fine-tuning on intermediate reasoning states. We keep the training procedure fixed and replace the confidence target with three alternative supervision signals. \textbf{Position} uses a target derived only from the position of the corresponding token in the reasoning trajectory, testing whether  learning reasoning progress is sufficient. \textbf{Binary correctness} replaces graded confidence with a binary target: $100\%$ if the trial answer is correct and $2\%$ otherwise. Unlike ConfSFT, this target requires access to the gold answer. Finally, \textbf{Shuffled confidence} randomly permutes the original confidence targets across training examples, preserving the label distribution while breaking the correspondence between each reasoning state and its confidence.

\begin{table}[b]
  \centering
  \caption{\textbf{Supervision-signal ablations.} Replacing graded confidence with position, binary correctness, or shuffled confidence yields weaker or less consistent efficiency gains.}
  \label{tab:ablations}
  \vspace{2pt}
  \setlength{\tabcolsep}{6pt}
  \renewcommand{\arraystretch}{1.12}
  \arrayrulecolor{tablerule}
  \small
  \newcommand{\ts}[2]{#1\,\textcolor{savedgreen}{\tiny(#2\%)}}
  \begin{tabular}{@{}lcccc@{}}
    \toprule
    & \multicolumn{2}{c}{Gemma-4-E2B} & \multicolumn{2}{c}{Nemotron-Nano-8B} \\
    \cmidrule(lr){2-3}\cmidrule(lr){4-5}
    Method & Acc.$\uparrow$ & Avg.\ Tokens$\downarrow$ & Acc.$\uparrow$ & Avg.\ Tokens$\downarrow$ \\
    \midrule
    Base & 48.7 & 7,065 & 60.0 & 9,996 \\
    Position & 46.5 & \ts{6,550}{-7.3} & 61.0 & \ts{10,122}{+1.3} \\
    Binary correctness & 48.5 & \ts{7,480}{+5.9} & 58.3 & \ts{9,703}{-2.9} \\
    Shuffled confidence & 48.7 & \ts{7,057}{-0.1} & 57.1 & \ts{10,392}{+4.0} \\
    \rowcolor{oursrow}
    \textbf{ConfSFT} & 48.3 & \ts{6,242}{-11.7} & 58.3 & \ts{9,025}{-9.7} \\
    \bottomrule
  \end{tabular}
\end{table}

Table~\ref{tab:ablations} shows that the supervision signal matters. Shuffling the confidence targets largely removes or reverses the efficiency gains, while position and binary-correctness supervision yield weaker and less consistent improvements across the two models. These results suggest that the gains are not simply due to fine-tuning on intermediate reasoning states or providing a generic notion of progress; preserving the correspondence between each state and its graded confidence target appears important. Accuracy and token reduction across training rounds are shown in Appendix~\ref{app:ablation-rounds}.

\subsection{Choice of Decision Points}


By default, ConfSFT identifies decision points using occurrences of the textual marker \texttt{Wait}. However, the method does not require this marker to occur frequently: Gemma-4-E2B produces only about four \texttt{Wait} occurrences per trajectory on average, compared with more than 20 for the other model families, yet still shows substantial efficiency gains in Table~\ref{tab:main}. This raises a broader question of whether \texttt{Wait} itself is important, or whether it simply provides a convenient way to sample intermediate reasoning states. To test this, we use paragraph boundaries (\texttt{\textbackslash n\textbackslash n}) as a more general marker, which all evaluated models emit consistently and more than 100 times per trajectory on average.

Table~\ref{tab:decision} shows that ConfSFT remains effective when decision points are identified using \texttt{\textbackslash n\textbackslash n}. On Gemma, this variant reduces generated tokens by 8.4\% while maintaining average accuracy, compared with a 10.3\% reduction using \texttt{Wait}. Because paragraph boundaries provide substantially more candidate decision points, we use fewer training problems per round and a smaller learning rate for this variant: 4 problems per round and $1\times10^{-6}$, compared with 80 problems and $2\times10^{-6}$ for \texttt{Wait}. Thus, the two rows should not be interpreted as a controlled comparison of the markers themselves; rather, they show that ConfSFT can remain effective with a generic structural marker under an appropriately tuned training configuration.

\begin{table}[t]
  \centering
  \caption{\textbf{Choice of decision-point marker on Gemma-4-E2B.}
  ConfSFT remains effective when \texttt{Wait} is replaced with paragraph breaks
  (\texttt{\textbackslash n\textbackslash n}).}
  \label{tab:decision}
  \vspace{2pt}
  \setlength{\tabcolsep}{3.2pt}
  \renewcommand{\arraystretch}{1.12}
  \scriptsize

  \newcommand{\ts}[2]{#1\,\textcolor{savedgreen}{\tiny(#2\%)}}
  \newcommand{\sv}[1]{\textcolor{savedgreen}{#1\%}}

  \resizebox{\linewidth}{!}{%
  \begin{tabular}{@{}lcccccccccccc@{}}
    \toprule
    & \multicolumn{4}{c}{\textsc{Math}}
    & \multicolumn{2}{c}{\textsc{Science}}
    & \multicolumn{4}{c}{\textsc{Coding}}
    & \multicolumn{2}{c}{\textsc{Average}} \\
    \cmidrule(lr){2-5}
    \cmidrule(lr){6-7}
    \cmidrule(lr){8-11}
    \cmidrule(lr){12-13}

    \textbf{Marker}
    & \multicolumn{2}{c}{AIME2025}
    & \multicolumn{2}{c}{GSM8K}
    & \multicolumn{2}{c}{GPQA-Diamond}
    & \multicolumn{2}{c}{LiveCodeBench}
    & \multicolumn{2}{c}{HumanEval}
    & Acc.$\uparrow$
    & Token Red.$\uparrow$ \\
    \cmidrule(lr){2-3}
    \cmidrule(lr){4-5}
    \cmidrule(lr){6-7}
    \cmidrule(lr){8-9}
    \cmidrule(lr){10-11}

    & Acc.$\uparrow$ & Avg.\ Tokens$\downarrow$
    & Acc.$\uparrow$ & Avg.\ Tokens$\downarrow$
    & Acc.$\uparrow$ & Avg.\ Tokens$\downarrow$
    & Acc.$\uparrow$ & Avg.\ Tokens$\downarrow$
    & Acc.$\uparrow$ & Avg.\ Tokens$\downarrow$
    & & \\
    \midrule

    Base
    & 35.0 & 7,284
    & 91.2 & 956
    & 42.6 & 3,294
    & 42.1 & 7,543
    & 93.2 & 2,210
    & 60.8 & \sv{0.0} \\

    \texttt{Wait}
    & 32.1 & \ts{5,987}{-17.8}
    & 91.4 & \ts{938}{-1.9}
    & 40.8 & \ts{2,968}{-9.9}
    & 42.4 & \ts{6,403}{-15.1}
    & 93.8 & \ts{2,059}{-6.8}
    & 60.1 & \sv{-10.3} \\

    \texttt{\textbackslash n\textbackslash n}
    & 36.3 & \ts{6,241}{-14.3}
    & 91.2 & \ts{935}{-2.2}
    & 40.9 & \ts{3,015}{-8.5}
    & 41.0 & \ts{6,741}{-10.6}
    & 93.0 & \ts{2,071}{-6.3}
    & 60.5 & \sv{-8.4} \\

    \bottomrule
  \end{tabular}%
  }
\end{table}

More generally, we view the textual marker as a mechanism for sampling intermediate reasoning states rather than as a semantically meaningful component of ConfSFT. The results with both \texttt{Wait} and \texttt{\textbackslash n\textbackslash n} are consistent with this view, suggesting that the method does not depend on a particular marker as long as it provides sufficient coverage of the reasoning trajectory.

\section{Conclusion}

We show that reasoning efficiency can emerge from self-supervised confidence training, even when the training objective does not explicitly encourage shorter reasoning or stopping and inference remains unchanged. By training models to predict confidence at intermediate reasoning states using only their own token probabilities, we obtain substantial reductions in generated tokens at matched accuracy across multiple model families and mathematical, scientific, and coding benchmarks, despite training on only 600 AIME problems. Confidence supervision achieves accuracy--efficiency tradeoffs comparable to methods that explicitly optimize for shorter reasoning. These results suggest that more efficient reasoning can emerge as a downstream consequence of learning confidence itself.

\textbf{Reproducibility.} We provide detailed descriptions of all components to facilitate reproducibility. Specifically, we provide our prompts in Appendix~\ref{app:prompts}, training configurations in Appendix~\ref{app:train}, evaluation protocol in Appendix~\ref{app:eval}, baseline implementations in Appendix~\ref{app:baselines}, and details of our Schoenfeld analysis in Appendix~\ref{app:schoenfeld}.

\section*{Acknowledgments}

This work was supported in part by NSF CAREER Award 1942230, the ONR PECASE Award N00014-25-1-2378, ARO Early Career Program Award 310902-00001, Army Grant W911NF-21-2-0076, NSF Award CCF-2212458, NSF Award 2229885 (NSF Institute for Trustworthy AI in Law and Society, TRAILS), MURI Grant 14262683, DARPA AIQ Grant HR00112590066, and a Meta Research Award 314593-00001
\bibliography{iclr2027_conference}
\bibliographystyle{iclr2027_conference}

\appendix
\section{Prompts and training sequences}
\label{app}
\label{app:prompts}

This section gives the exact prompts and training-example construction used
by ConfSFT. We use the notation of Section~\ref{sec:method}.

\subsection{Reasoning prompt (rollouts and evaluation)}
\label{app:prompts:reasoning}

Training rollouts and the mathematical evaluations use the same user prompt,
applied through each model's chat template with the generation prompt
appended. No confidence instruction is added at training or at test time.
GPQA-Diamond, LiveCodeBench, and HumanEval replace the boxed line with the
benchmark-specific instructions in Appendix~\ref{app:eval:prompts}.

\begin{verbatim}
{problem}

Please reason step by step, and put your final answer within \boxed{}.
\end{verbatim}

The system prompt is \texttt{detailed thinking on} for Nemotron, which is
how that family enables chain-of-thought, and empty for Gemma and Qwen.
Gemma and Qwen are rendered with \texttt{enable\_thinking=True} so the
chat template opens a thinking block. GPT-OSS-20B uses the Harmony chat
template with reasoning effort \texttt{medium} and no additional system
prompt.

\subsection{Decision-point markers}
\label{app:prompts:anchors}

By default, decision points are occurrences of \texttt{Wait} within the
model's reasoning. We identify them with the regular expression
\verb|\bWait\b| and retain only matches before the end-of-thinking marker
(\verb+</think>+ for Nemotron and Qwen; \verb+<channel|>+ for Gemma;
\verb+<|channel|>final+ for GPT-OSS-20B).
When a trajectory contains more than $M=32$ decision points, we retain
$32$ approximately uniformly spaced points in their original order.
For Gemma, special tokens are retained while decoding so that
\verb+<channel|>+ remains available as the end-of-thinking marker.
GPT-OSS-20B does the same for \verb+<|channel|>final+.

The paragraph-boundary variant in Section~\ref{sec:ablations} uses the same
rule with the marker \texttt{\textbackslash n\textbackslash n} in place of
\texttt{Wait}.

\subsection{Trial-answer probe}
\label{app:prompts:probe}

At a decision point, we append the following answer-elicitation prompt $q$
to the problem prompt and reasoning prefix $y_{<t}$. The decision-point
marker itself is excluded from the prefix.

\begin{verbatim}
\n**Final Answer**\n\nThe final answer is \boxed{
\end{verbatim}

We greedily generate at most $16$ tokens and stop at a closing brace,
newline, end-of-thinking marker, or end-of-sequence token. A terminating
stop token is excluded from the score. The trial answer $a$ is the
brace-balanced span following the open brace in $q$, and $c_t$ is the
geometric mean of its token probabilities, as defined in
Equation~\ref{eq:confidence}. Constructing the ConfSFT target never uses
the gold answer.

\subsection{Priming prefix and label}
\label{app:prompts:prefix}

We use $K=50$ percentage levels in $2\%$ increments. Each
$c_t\in[0,1]$ is rounded up to the smallest grid point
$\ell_t\in\{2\%,4\%,\dots,100\%\}$ that is at least $c_t$.
The priming prefix $p$ is the sentence used in Section~\ref{sec:method}:

\begin{verbatim}
From 0% (very low) to 100% (very high), my confidence in the answer so far is
\end{verbatim}

\subsection{Training sequence and loss mask}
\label{app:prompts:sequence}

A training example concatenates the chat prompt, the reasoning prefix
$y_{<t}$, the priming prefix $p$ and the label $\ell_t$ (Equation~\ref{eq:sequence}). A single space separates the reasoning from
$p$ when the prefix does not already end in whitespace, and a single space
separates $p$ from $\ell_t$. The tail of such a sequence reads

\begin{verbatim}
... so the exponents are {10,9,8,7,6,5,3}. From 0% (very low) to 100%
(very high), my confidence in the answer so far is 72%
\end{verbatim}

and the loss mask over its segments is

\begin{verbatim}
segment                                          supervised
-----------------------------------------------  ----------
chat prompt (system + problem + instruction)     no
reasoning prefix y_<t>                           no
priming prefix p                                 no
separating space                                 no
"72%"                                            yes
\end{verbatim}

Only the tokens of $\ell_t$ receive a loss; every other position, including
padding, is assigned the ignore index and does not contribute to
Equation~\ref{eq:loss}. We use a maximum training context of $16{,}384$
tokens for Nemotron, Gemma, and Qwen. For GPT-OSS-20B the limit is
$8{,}192$ tokens, for both the supervised sequence and the generated
rollout, because a $16{,}384$-token update does not fit our compute budget
on this model (Appendix~\ref{app:train:gpt}).

\section{Training hyperparameters}
\label{app:train}

\subsection{Shared recipe}
\label{app:train:shared}

We use full-parameter fine-tuning in \texttt{bfloat16} with FSDP for
Nemotron, Gemma, and Qwen. GPT-OSS-20B is trained with LoRA instead
(Appendix~\ref{app:train:gpt}). We use
Adafactor with learning rates given in Table~\ref{tab:app-hparams}, no
weight decay, gradient clipping at $1.0$, and linear warmup over the first
$3\%$ of optimizer steps in each round. The per-device batch size is $1$
with $4$ gradient-accumulation steps. We make one pass over the confidence
examples constructed in each round.

Training uses the AIME problems from 2000--2023
\citep{aime19832024}, while the $30$ AIME~2024 problems are held out for
validation and checkpoint selection. The training set contains $695$
problems. We shuffle it once with seed $42$ and divide it into eight
disjoint groups: seven groups of $87$ problems and one group of $86$.
At round $r$, the current policy $\pi_{\theta_r}$ generates $8$ rollouts
for each problem in group $r$. We construct one confidence-supervision
example for each retained decision point and fine-tune on the resulting
examples to obtain $\pi_{\theta_{r+1}}$. After each round, we generate
$16$ validation rollouts per AIME~2024 problem. Training and validation
rollouts use the boxed prompt and model-specific decoding parameters in
Appendix~\ref{app:eval:models}, with at most $16{,}384$ new tokens
($8{,}192$ for GPT-OSS-20B).

\subsection{Per-model settings}
\label{app:train:models}

Table~\ref{tab:app-hparams} lists the settings that differ across the three
ConfSFT models in Table~\ref{tab:main}. The Hugging Face checkpoints are
\texttt{nvidia/Llama-3.1-Nemotron-Nano-8B-v1},
\texttt{google/gemma-4-E2B-it}, and \texttt{Qwen/Qwen3-4B}.
Attention is SDPA for Nemotron and Qwen and eager attention for Gemma.
Gemma additionally keeps special tokens at decode time, as noted above.

\begin{table}[t]
  \centering
  \caption{\textbf{ConfSFT hyperparameters} for the three models in
  Table~\ref{tab:main}. Shared settings are in the text above.}
  \label{tab:app-hparams}
  \small
  \setlength{\tabcolsep}{4pt}
  \begin{tabular}{@{}l ccc@{}}
    \toprule
    & Nemotron-Nano-8B & Gemma-4-E2B & Qwen3-4B \\
    \midrule
    System prompt & \texttt{detailed thinking on} & (none) & (none) \\
    Think-end marker & \verb+</think>+ & \verb+<channel|>+ & \verb+</think>+ \\
    $T$ / top-$p$ / top-$k$
      & $0.6$ / $0.95$ / $20$
      & $1.0$ / $0.95$ / $64$
      & $0.6$ / $0.95$ / $20$ \\
    Learning rate & $1\times 10^{-6}$ & $2\times 10^{-6}$ & $1\times 10^{-6}$ \\
    Groups / problems per round & $8$ / $87$ & $8$ / $87$ & $8$ / $87$ \\
    Rollouts / training problem & $8$ & $8$ & $8$ \\
    Validation rollouts / problem & $16$ & $16$ & $16$ \\
    Decision-point marker & \texttt{Wait} & \texttt{Wait} & \texttt{Wait} \\
    Selected round & $4$ & $3$ & $4$ \\
    \bottomrule
  \end{tabular}
\end{table}

\subsection{GPT-OSS-20B}
\label{app:train:gpt}

GPT-OSS-20B (\texttt{openai/gpt-oss-20b}) is a mixture-of-experts model.
Its expert weights are stored in MXFP4, and a full-parameter update does
not fit our training setup, so we fine-tune a LoRA adapter and leave the
base weights frozen, including every expert layer. The adapter has rank
$r=16$, scaling $\alpha=32$, and dropout $0$, and is applied only to the
attention projections \texttt{q\_proj}, \texttt{k\_proj}, \texttt{v\_proj},
and \texttt{o\_proj}. For the supervised step we dequantize the MXFP4
weights to \texttt{bfloat16}. Generation keeps the native MXFP4 weights
and applies the adapter on top.

The optimizer, warmup, clipping, and loss mask match
Section~\ref{app:train:shared}. The learning rate is $2\times 10^{-5}$,
the per-device batch size is $1$, and we do not accumulate gradients.
We train for one round on the first group of $87$ problems, with $2$
rollouts per problem, and report that round-$1$ adapter. Decision points are
\texttt{Wait}, and the end-of-thinking marker is
\verb+<|channel|>final+. Reasoning effort is \texttt{medium}.

\subsection{Ablation and decision-point variants}
\label{app:train:ablations}

For the supervision-signal ablations in Table~\ref{tab:ablations}, we keep
the training construction fixed and replace only the target that is
expressed as $\ell_t$. \textbf{Position} is the fraction of thinking
tokens preceding the decision point. \textbf{Binary correctness} is
$100\%$ when the trial answer is correct and $2\%$ otherwise.
\textbf{Shuffled confidence} randomly permutes the ConfSFT labels across
training examples, preserving their marginal distribution.

The Nemotron ablations use the main Nemotron configuration and learning
rate $1\times10^{-6}$. We report round~$4$ for ConfSFT, Position, and
Binary correctness, and round~$2$ for Shuffled confidence. The Gemma
ablations use paragraph boundaries as decision points, four training
problems per round, learning rate $1\times10^{-6}$, and the round-$5$
checkpoint. This is also the paragraph-boundary configuration reported in
Table~\ref{tab:decision}.

\section{Evaluation protocol}
\label{app:eval}

Table~\ref{tab:main} uses standard generation with no confidence instruction, answer cue, or priming prefix. For each problem, we sample 16 independent completions. We report accuracy as mean correctness across all samples (\textbf{pass@1}) and \textbf{pass@8} using the unbiased estimator
\[
\mathrm{pass@}k
=
1-\frac{\binom{n-c}{k}}{\binom{n}{k}},
\]
where $n=16$ is the number of sampled completions and $c$ is the number of correct completions for the problem. This corresponds to the probability that at least one correct answer appears among $k$ samples.

\textbf{Avg.\ Tokens} is the mean number of generated completion tokens. \textbf{Token Red.} is computed relative to the corresponding base model as
\[
100\times\left(1-\frac{t_{\mathrm{method}}}{t_{\mathrm{base}}}\right).
\]

\subsection{Models and decoding}
\label{app:eval:models}

Decoding follows each model family's recommended sampler:

\begin{center}
\small
\begin{tabular}{@{}l l ccc@{}}
  \toprule
  Model & Mode & $T$ & top-$p$ & top-$k$ \\
  \midrule
  Gemma-4-E2B & \texttt{enable\_thinking=True} & $1.0$ & $0.95$ & $64$ \\
  Nemotron-Nano-8B & \texttt{detailed thinking on} & $0.6$ & $0.95$ & $20$ \\
  Qwen3-4B & \texttt{enable\_thinking=True} & $0.6$ & $0.95$ & $20$ \\
  GPT-OSS-20B & \texttt{Reasoning: medium} & $1.0$ & $1.0$ & $40$ \\
  \bottomrule
\end{tabular}
\end{center}

We use a maximum of $8{,}192$ generated tokens for GPT-OSS-20B, matching the context used during confidence training, and $16{,}384$ for the other models. For LiveCodeBench, we increase the limit to $32{,}768$.

\subsection{Benchmarks and sample counts}
\label{app:eval:benches}

We use the Hugging Face versions listed below. For LiveCodeBench, we use the lite code-generation split from releases v5 and v6 and keep problems with \texttt{contest\_date} $\ge$ $2025$-$01$-$01$, resulting in 182 problems from contests between $2025$-$01$-$04$ and $2025$-$04$-$06$. For GPQA-Diamond, we shuffle the answer choices once per question and use the same ordering for all methods.

\begin{center}
\small
\begin{tabular}{@{}l l r r@{}}
  \toprule
  Benchmark & Hugging Face source & Problems & Samples / problem \\
  \midrule
  AIME 2025 & \texttt{opencompass/AIME2025} (parts I+II) & $30$ & $16$ \\
  GSM8K & \texttt{openai/gsm8k} (\texttt{main} test) & $1{,}319$ & $16$ \\
  GPQA-Diamond & \texttt{Idavidrein/gpqa} (\texttt{gpqa\_diamond}) & $198$ & $16$ \\
  LiveCodeBench & \texttt{livecodebench/code\_generation\_lite} (v5+v6)
    & $182$ & $16$ \\
  HumanEval & \texttt{openai/openai\_humaneval} (test) & $164$ & $16$ \\
  \bottomrule
\end{tabular}
\end{center}

\subsection{Evaluation prompts}
\label{app:eval:prompts}

AIME 2025 and GSM8K use the boxed instruction of
Appendix~\ref{app:prompts:reasoning}. GPQA-Diamond asks for a letter in a
box. Code benchmarks ask for a markdown fence rather than a box, and
LiveCodeBench keeps the official stdin versus starter-code wording.

\begin{verbatim}
GPQA-Diamond:
{question}

(A) ...
(B) ...
(C) ...
(D) ...

Reason step by step and put your final answer letter (A, B, C, or D) in \boxed{}.

LiveCodeBench (no starter code):
{question}

Read the inputs from stdin, solve the problem, and write the answer to stdout
(do not directly test on the sample inputs). Enclose your code within delimiters
as follows.

Please reason step by step. Put your final Python solution in a markdown code
block: ```python ... ```.

LiveCodeBench (with starter code):
{question}

You will use the following starter code to write the solution to the problem
and enclose your code within delimiters.
```python
{starter_code}
```

Please reason step by step. Put your final Python solution in a markdown code
block: ```python ... ```.

HumanEval:
Complete the following Python function. Read the docstring carefully.

```python
{function_prompt}
```

Please reason step by step. Put your final Python solution in a markdown code
block: ```python ... ```. The block must define the function described above
(including the signature).
\end{verbatim}

\subsection{Answer extraction and grading}
\label{app:eval:grade}

We use the same fixed extraction and grading procedure for the base model and all evaluated methods. Although the prompts specify a final-answer format, models occasionally produce the correct answer without following it exactly. We therefore use simple fallback rules to recover answers or code already present in the completion rather than marking these cases incorrect.

\paragraph{AIME 2025 and GSM8K.}
We first look for an answer in \verb|\boxed{...}| and take the first number inside the last box. If no boxed answer is present, we take the last number appearing in the final $160$ characters of the completion. The extracted number is compared with the first number in the gold answer, so, for example, a prediction of $336$ is credited against gold $336^\circ$.

\paragraph{GPQA-Diamond.}
We first extract the final answer letter A--D from the last \verb|\boxed{...}|. If no boxed answer is present, we take the last standalone letter A--D in the final $160$ characters. The extracted letter is compared with the gold answer after applying the fixed answer-choice ordering used for that question.

\paragraph{LiveCodeBench.}
We extract Python code from the completion and evaluate it with the official LiveCodeBench tests using a $6$s timeout. We first use the last non-empty closed markdown code block. If the model opens a final code block but does not close it, we also consider the text following that fence. When the benchmark expects a \texttt{Solution} method but the generated code instead defines the corresponding standalone function, we additionally wrap that function in the required interface. A sample is marked correct if one of these extracted versions passes all tests. These fallback rules only recover code already present in the completion or adapt its interface; we do not repair or otherwise modify the generated solution.

\paragraph{HumanEval.}
We use the last markdown code block, dropping a leading language tag when present. If no usable code block is found, we use the code produced after the model's reasoning section. If the extracted program already defines the target function, we execute it directly; otherwise, we treat it as a continuation of the dataset prompt and prepend the original function prompt. We evaluate with the official HumanEval tests using a $3$s timeout.

\subsection{More evaluation results}
\label{app:ci}
Table~\ref{tab:main-ci} reports paired question-level 95\% confidence intervals for the results in Table~\ref{tab:main}. Each method is compared with its corresponding base model on the same set of questions. Accuracy values are differences in percentage points, while token values are differences in average generated tokens. The intervals show that the token reductions from ConfSFT are consistent across model families, while accuracy remains close to the corresponding base models.

Table~\ref{tab:main-pass8} additionally reports pass@8, computed with the unbiased estimator described in Appendix~\ref{app:eval}. The overall pattern is consistent with pass@1: ConfSFT substantially reduces generated tokens while preserving pass@8 close to the corresponding base model across all four model families.

\begin{table}[t]
  \centering
  \caption{\textbf{Main results relative to the base model,} with paired
  $95\%$ CIs. Acc and tokens are differences from the
  corresponding base row of Table~\ref{tab:main}.}
  \label{tab:main-ci}
  \vspace{2pt}
  \setlength{\tabcolsep}{2.2pt}
  \renewcommand{\arraystretch}{1.12}
  \arrayrulecolor{tablerule}
  \scriptsize
  \newcommand{\hacc}{\multicolumn{1}{c}{Acc$\uparrow$}}
  \newcommand{\htok}{\multicolumn{1}{c}{Avg.\ Tokens$\downarrow$}}
  \newcommand{\hsav}{\multicolumn{1}{c}{Token Red.}}
  \newcommand{\haccG}{\multicolumn{1}{!{\color{tablerule}\vrule}c}{Acc$\uparrow$}}
  \newcommand{\dom}[2]{\multicolumn{#1}{c!{\color{tablerule}\vrule}}{\textsc{#2}}}
  \newcommand{\domend}[2]{\multicolumn{#1}{c}{\textsc{#2}}}
  \newcommand{\bn}[1]{\multicolumn{2}{c!{\color{tablerule}\vrule}}{#1}}
  \newcommand{\bnend}[1]{\multicolumn{2}{c}{#1}}
  \newcommand{\ts}[2]{#1\,\textcolor{savedgreen}{\tiny(#2\%)}}
  \newcommand{\sv}[1]{\textcolor{savedgreen}{\ensuremath{#1}\%}}
  \resizebox{\linewidth}{!}{%
  \begin{tabular}{@{}l !{\color{tablerule}\vrule}
      c@{\hspace{4pt}}c !{\color{tablerule}\vrule}
      c@{\hspace{4pt}}c !{\color{tablerule}\vrule}
      c@{\hspace{4pt}}c !{\color{tablerule}\vrule}
      c@{\hspace{4pt}}c !{\color{tablerule}\vrule}
      c@{\hspace{4pt}}c !{\color{tablerule}\vrule}
      c@{\hspace{4pt}}c@{}}
    \toprule
    \multirow{3}{*}{Method}
    & \dom{4}{Math} & \dom{2}{Science} & \dom{4}{Coding} & \domend{2}{Average} \\
    & \bn{AIME2025} & \bn{GSM8K} & \bn{GPQA-Diamond} & \bn{LiveCodeBench} & \bn{HumanEval} & \bnend{} \\
    \cmidrule(lr){2-3}\cmidrule(lr){4-5}\cmidrule(lr){6-7}\cmidrule(lr){8-9}\cmidrule(lr){10-11}\cmidrule(lr){12-13}
    & \hacc & \htok & \haccG & \htok
           & \haccG & \htok & \haccG & \htok
           & \haccG & \htok & \haccG & \hsav \\
    \midrule
    \multicolumn{13}{@{}l}{\textit{Nemotron-Nano-8B}} \\
    DEER\citep{deer}  & $-2.7{\pm}2.3$ & $-2{,}249{\pm}593$ & $-2.6{\pm}0.5$ & $-427{\pm}29$ & $-2.7{\pm}1.5$ & $-1{,}626{\pm}220$ & $-26.8{\pm}4.4$ & $-10{,}030{\pm}944$ & $-42.6{\pm}4.7$ & $-3{,}089{\pm}701$ & $-15.5{\pm}1.4$ & \sv{-49.8{\pm}1.6} \\
    On-Policy SFT\citep{onpolicysft} & $-3.5{\pm}4.9$ & $-838{\pm}351$ & $-0.7{\pm}0.5$ & $-193{\pm}25$ & $-2.2{\pm}2.1$ & $-1{,}192{\pm}248$ & $-2.3{\pm}1.4$ & $-397{\pm}189$ & $-0.4{\pm}1.2$ & $-426{\pm}234$ & $-1.8{\pm}1.1$ & \sv{-11.0{\pm}1.6} \\
    A\&Z\citep{arora2025traininglanguagemodelsreason} & $-1.0{\pm}5.4$ & $-1{,}681{\pm}458$ & $-0.8{\pm}0.3$ & $+26{\pm}16$ & $-1.6{\pm}1.8$ & $-600{\pm}151$ & $+0.3{\pm}1.3$ & $-415{\pm}224$ & $-1.3{\pm}1.2$ & $-112{\pm}253$ & $-0.9{\pm}1.2$ & \sv{-5.5{\pm}1.6} \\
    \rowcolor{oursrow}
    \textbf{ConfSFT (ours)} & $+3.1{\pm}4.6$ & $-1{,}173{\pm}385$ & $-0.2{\pm}0.3$ & $-182{\pm}17$ & $+0.3{\pm}1.7$ & $-1{,}060{\pm}128$ & $+1.0{\pm}1.1$ & $-610{\pm}191$ & $+1.1{\pm}1.3$ & $-376{\pm}275$ & $+1.1{\pm}1.0$ & \sv{-11.1{\pm}1.6} \\
    \midrule
\multicolumn{13}{@{}l}{\textit{Gemma-4-E2B}} \\

DEER\citep{deer}
& $-10.8{\pm}5.6$ & $-2{,}360{\pm}630$
& $-0.8{\pm}0.2$ & $-44{\pm}6$
& $-6.1{\pm}1.9$ & $-862{\pm}125$
& $-10.8{\pm}2.4$ & $-2{,}210{\pm}310$
& $-10.3{\pm}2.2$ & $-305{\pm}89$
& $-7.8{\pm}1.4$ & \sv{-21.3{\pm}1.9} \\

On-Policy SFT\citep{onpolicysft}
& $-7.7{\pm}5.6$ & $-1{,}684{\pm}538$
& $-0.3{\pm}0.4$ & $+93{\pm}6$
& $-0.5{\pm}1.9$ & $+293{\pm}54$
& $-1.1{\pm}1.4$ & $+168{\pm}103$
& $0.0{\pm}1.5$ & $+108{\pm}32$
& $-1.9{\pm}1.3$ & \sv{+0.5{\pm}1.4} \\

A\&Z\citep{arora2025traininglanguagemodelsreason}
& $+0.4{\pm}3.9$ & $-266{\pm}171$
& $+0.1{\pm}0.3$ & $+7{\pm}5$
& $-1.1{\pm}1.5$ & $-70{\pm}31$
& $-0.9{\pm}1.6$ & $-252{\pm}96$
& $+0.3{\pm}1.2$ & $-10{\pm}34$
& $-0.2{\pm}0.9$ & \sv{-1.8{\pm}0.6} \\

\rowcolor{oursrow}
\textbf{ConfSFT (ours)}
& $-2.9{\pm}3.0$ & $-1{,}297{\pm}244$
& $+0.2{\pm}0.3$ & $-18{\pm}4$
& $-1.8{\pm}1.7$ & $-326{\pm}44$
& $+0.3{\pm}1.3$ & $-1{,}139{\pm}129$
& $+0.6{\pm}1.0$ & $-151{\pm}39$
& $-0.7{\pm}0.8$ & \sv{-10.3{\pm}0.7} \\
    \midrule
    \multicolumn{13}{@{}l}{\textit{Qwen3-4B}} \\
    DEER\citep{deer}  & $-18.8{\pm}9.8$ & $-4{,}182{\pm}1121$ & $-9.9{\pm}1.8$ & $-1{,}711{\pm}84$ & $-6.0{\pm}4.0$ & $-5{,}455{\pm}427$ & $-21.1{\pm}3.9$ & $-4{,}580{\pm}580$ & $-42.5{\pm}4.8$ & $-2{,}738{\pm}524$ & $-19.6{\pm}2.5$ & \sv{-55.0{\pm}2.4} \\
    On-Policy SFT\citep{onpolicysft} & $-0.4{\pm}7.3$ & $-1{,}863{\pm}665$ & $+0.2{\pm}0.6$ & $-877{\pm}56$ & $+0.9{\pm}3.6$ & $-1{,}381{\pm}180$ & $-0.9{\pm}1.5$ & $-436{\pm}322$ & $-0.1{\pm}0.8$ & $-550{\pm}169$ & $-0.1{\pm}1.7$ & \sv{-17.2{\pm}1.7} \\
    \rowcolor{oursrow}
    \textbf{ConfSFT (ours)} & $-0.4{\pm}5.2$ & $-1{,}903{\pm}583$ & $-0.3{\pm}0.8$ & $-583{\pm}63$ & $+2.7{\pm}2.9$ & $-2{,}237{\pm}215$ & $-1.9{\pm}1.3$ & $-2{,}832{\pm}383$ & $0.0{\pm}1.1$ & $-440{\pm}207$ & $0.0{\pm}1.2$ & \sv{-19.2{\pm}1.6} \\
    \midrule
    \multicolumn{13}{@{}l}{\textit{gpt-oss-20b}} \\
    DEER\citep{deer}  & $0.0{\pm}0.0$ & $+5{\pm}24$ & $0.0{\pm}0.0$ & $+1{\pm}0$ & $0.0{\pm}0.0$ & $+21{\pm}4$ & $-2.5{\pm}0.8$ & $-209{\pm}91$ & $-5.3{\pm}1.9$ & $-84{\pm}45$ & $-1.6{\pm}0.4$ & \sv{-2.5{\pm}0.9} \\
    \rowcolor{oursrow}
    \textbf{ConfSFT (ours)} & $+0.6{\pm}4.3$ & $-567{\pm}187$ & $+0.1{\pm}0.3$ & $-44{\pm}9$ & $-0.1{\pm}1.6$ & $-620{\pm}89$ & $-2.9{\pm}2.3$ & $-641{\pm}110$ & $-0.9{\pm}1.4$ & $-66{\pm}25$ & $-0.6{\pm}1.1$ & \sv{-11.1{\pm}1.0} \\
    \bottomrule
  \end{tabular}%
  }
\end{table}

\begin{table}[t]
  \centering
  \caption{\textbf{Pass@8.} Same methods and benchmarks as Table~\ref{tab:main}.}
  \label{tab:main-pass8}
  \vspace{2pt}
  \setlength{\tabcolsep}{2.2pt}
  \renewcommand{\arraystretch}{1.12}
  \arrayrulecolor{tablerule}
  \scriptsize
  \newcommand{\hacc}{\multicolumn{1}{c}{P@8$\uparrow$}}
  \newcommand{\htok}{\multicolumn{1}{c}{Avg.\ Tokens$\downarrow$}}
  \newcommand{\hsav}{\multicolumn{1}{c}{Token Red.}}
  \newcommand{\haccG}{\multicolumn{1}{!{\color{tablerule}\vrule}c}{P@8$\uparrow$}}
  \newcommand{\dom}[2]{\multicolumn{#1}{c!{\color{tablerule}\vrule}}{\textsc{#2}}}
  \newcommand{\domend}[2]{\multicolumn{#1}{c}{\textsc{#2}}}
  \newcommand{\bn}[1]{\multicolumn{2}{c!{\color{tablerule}\vrule}}{#1}}
  \newcommand{\bnend}[1]{\multicolumn{2}{c}{#1}}
  \newcommand{\ts}[2]{#1\,\textcolor{savedgreen}{\tiny(#2\%)}}
  \newcommand{\sv}[1]{\textcolor{savedgreen}{#1\%}}
  \resizebox{\linewidth}{!}{%
  \begin{tabular}{@{}l !{\color{tablerule}\vrule}
      c@{\hspace{4pt}}c !{\color{tablerule}\vrule}
      c@{\hspace{4pt}}c !{\color{tablerule}\vrule}
      c@{\hspace{4pt}}c !{\color{tablerule}\vrule}
      c@{\hspace{4pt}}c !{\color{tablerule}\vrule}
      c@{\hspace{4pt}}c !{\color{tablerule}\vrule}
      c@{\hspace{4pt}}c@{}}
    \toprule
    \multirow{3}{*}{Method}
    & \dom{4}{Math} & \dom{2}{Science} & \dom{4}{Coding} & \domend{2}{Average} \\
    & \bn{AIME2025} & \bn{GSM8K} & \bn{GPQA-Diamond} & \bn{LiveCodeBench} & \bn{HumanEval} & \bnend{} \\
    \cmidrule(lr){2-3}\cmidrule(lr){4-5}\cmidrule(lr){6-7}\cmidrule(lr){8-9}\cmidrule(lr){10-11}\cmidrule(lr){12-13}
    & \hacc & \htok & \haccG & \htok
           & \haccG & \htok & \haccG & \htok
           & \haccG & \htok & \haccG & \hsav \\
    \midrule
    \multicolumn{13}{@{}l}{\textit{Nemotron-Nano-8B}} \\
    Base  & 67.1 & 11,325 & 97.0 & 1,192 & 82.3 & 7,338 & 59.4 & 11,761 & 97.9 & 3,600 & 80.7 & \sv{0.0} \\
    DEER\citep{deer}  & 66.9 & \ts{9,076}{-19.9} & 96.8 & \ts{765}{-35.8} & 79.7 & \ts{5,712}{-22.2} & 34.3 & \ts{1,731}{-85.3} & 80.5 & \ts{511}{-85.8} & 71.7 & \sv{-49.8} \\
    On-Policy SFT\citep{onpolicysft} & 67.7 & \ts{10,487}{-7.4} & 96.3 & \ts{999}{-16.2} & 81.9 & \ts{6,146}{-16.2} & 58.7 & \ts{11,364}{-3.4} & 97.5 & \ts{3,174}{-11.8} & 80.4 & \sv{-11.0} \\
    A\&Z\citep{arora2025traininglanguagemodelsreason} & 71.6 & \ts{9,644}{-14.8} & 96.8 & \ts{1,218}{+2.2} & 80.1 & \ts{6,738}{-8.2} & 61.1 & \ts{11,346}{-3.5} & 97.3 & \ts{3,489}{-3.1} & 81.4 & \sv{-5.5} \\
    \rowcolor{oursrow}
    \textbf{ConfSFT (ours)} & 69.7 & \ts{10,153}{-10.4} & 97.0 & \ts{1,010}{-15.3} & 81.9 & \ts{6,278}{-14.4} & 59.0 & \ts{11,151}{-5.2} & 98.3 & \ts{3,225}{-10.4} & 81.2 & \sv{-11.1} \\
    \midrule
    \multicolumn{13}{@{}l}{\textit{Gemma-4-E2B}} \\
    Base  & 56.9 & 7,284 & 95.8 & 956 & 75.3 & 3,294 & 57.7 & 7,543 & 98.6 & 2,210 & 76.9 & \sv{0.0} \\
    DEER\citep{deer}  & 46.5 & \ts{4,924}{-32.4} & 95.7 & \ts{912}{-4.6} & 70.9 & \ts{2,432}{-26.2} & 51.8 & \ts{5,333}{-29.3} & 97.7 & \ts{1,905}{-13.8} & 72.5 & \sv{-21.3} \\
    On-Policy SFT\citep{onpolicysft} & 45.5 & \ts{5,601}{-23.1} & 96.0 & \ts{1,049}{+9.7} & 75.0 & \ts{3,587}{+8.9} & 58.6 & \ts{7,711}{+2.2} & 99.4 & \ts{2,318}{+4.9} & 74.9 & \sv{+0.5} \\
    A\&Z\citep{arora2025traininglanguagemodelsreason} & 53.5 & \ts{7,018}{-3.7} & 96.0 & \ts{963}{+0.7} & 73.4 & \ts{3,224}{-2.1} & 55.5 & \ts{7,291}{-3.3} & 99.0 & \ts{2,200}{-0.5} & 75.5 & \sv{-1.8} \\
    \rowcolor{oursrow}
    \textbf{ConfSFT (ours)} & 57.3 & \ts{5,987}{-17.8} & 95.8 & \ts{938}{-1.9} & 73.6 & \ts{2,968}{-9.9} & 58.9 & \ts{6,403}{-15.1} & 99.1 & \ts{2,059}{-6.8} & 76.9 & \sv{-10.3} \\
    \midrule
    \multicolumn{13}{@{}l}{\textit{Qwen3-4B}} \\
    Base  & 73.3 & 13,268 & 94.8 & 2,292 & 72.2 & 9,075 & 58.9 & 14,954 & 97.1 & 3,495 & 79.3 & \sv{0.0} \\
    DEER\citep{deer}  & 73.3 & \ts{9,086}{-31.5} & 84.8 & \ts{580}{-74.7} & 64.1 & \ts{3,621}{-60.1} & 48.4 & \ts{10,374}{-30.6} & 85.4 & \ts{756}{-78.4} & 71.2 & \sv{-55.0} \\
    On-Policy SFT\citep{onpolicysft} & 76.7 & \ts{11,405}{-14.0} & 96.9 & \ts{1,414}{-38.3} & 77.1 & \ts{7,694}{-15.2} & 60.8 & \ts{14,518}{-2.9} & 97.2 & \ts{2,944}{-15.7} & 81.7 & \sv{-17.2} \\
    \rowcolor{oursrow}
    \textbf{ConfSFT (ours)} & 76.7 & \ts{11,365}{-14.3} & 94.5 & \ts{1,710}{-25.4} & 70.7 & \ts{6,838}{-24.6} & 56.1 & \ts{12,122}{-18.9} & 97.0 & \ts{3,054}{-12.6} & 79.0 & \sv{-19.2} \\
    \midrule
    \multicolumn{13}{@{}l}{\textit{gpt-oss-20b}} \\
    Base  & 78.4 & 5,802 & 97.7 & 482 & 85.8 & 3,732 & 67.4 & 5,034 & 99.3 & 917 & 85.7 & \sv{0.0} \\
    DEER\citep{deer}  & 78.4 & \ts{5,807}{+0.1} & 97.7 & \ts{482}{+0.1} & 85.8 & \ts{3,753}{+0.6} & 67.0 & \ts{4,824}{-4.2} & 99.1 & \ts{832}{-9.2} & 85.6 & \sv{-2.5} \\
    \rowcolor{oursrow}
    \textbf{ConfSFT (ours)} & 76.3 & \ts{5,235}{-9.8} & 97.8 & \ts{437}{-9.2} & 88.3 & \ts{3,112}{-16.6} & 66.1 & \ts{4,393}{-12.7} & 98.8 & \ts{850}{-7.2} & 85.5 & \sv{-11.1} \\
    \bottomrule
  \end{tabular}%
  }
\end{table}

\section{Additional analyses on Gemma-4-E2B}
\label{app:cmaegemma}
\label{app:gemma-analysis}

\begin{figure}[t]
  \centering
  \includegraphics[width=\linewidth]{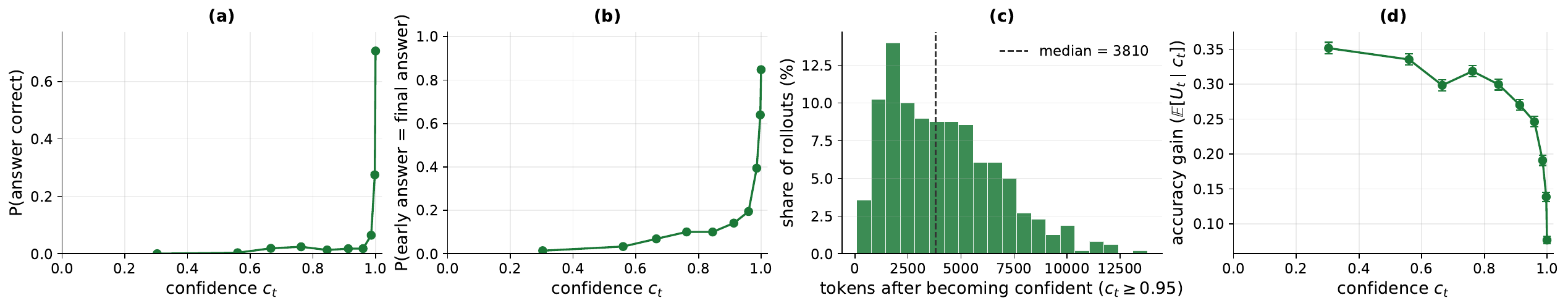}
  \caption{Gemma-4-E2B counterpart of Figure~\ref{fig:conf-waste-four}.}
  \label{fig:conf-waste-four-gemma}
\end{figure}

Figure~\ref{fig:conf-waste-four-gemma} repeats the four-panel analysis of
Figure~\ref{fig:conf-waste-four} on the Gemma-4-E2B base model.

We repeat the confidence-learning analyses from
Section~\ref{sec:results} on Gemma-4-E2B. 

\begin{figure}[t]
  \centering
  \includegraphics[width=\linewidth]{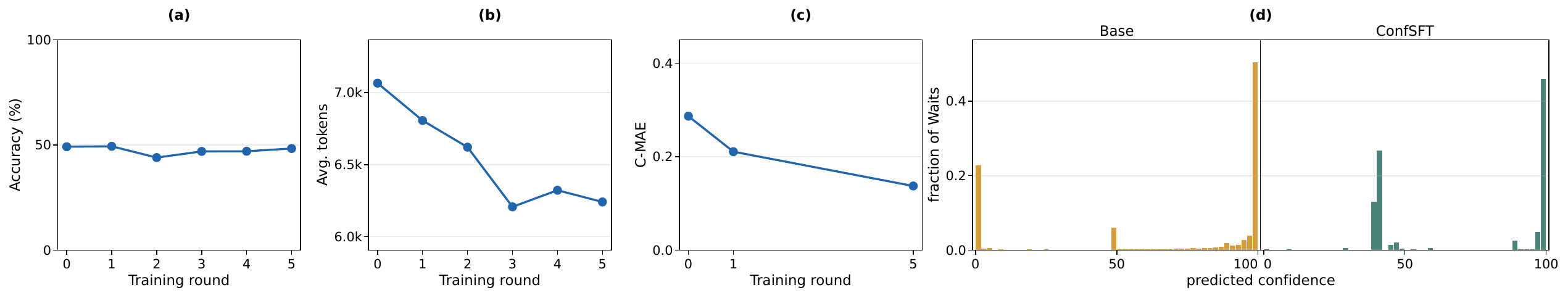}
  \caption{Gemma-4-E2B counterpart of Figure~\ref{fig:cmae-over-epochs}.}
  \label{fig:cmae-over-epochs-gemma}
\end{figure}

Figure~\ref{fig:cmae-over-epochs-gemma} repeat the confidence-learning analyses from
Section~\ref{sec:results} on Gemma-4-E2B. It tracks C-MAE, tokens,
and accuracy across training rounds.
C-MAE and tokens falls over training, and accuracy
stays within a few points of the base model.

\section{Supervision-signal ablations over training rounds}
\label{app:ablation-rounds}

\begin{figure}[t]
  \centering
  \includegraphics[width=\linewidth]{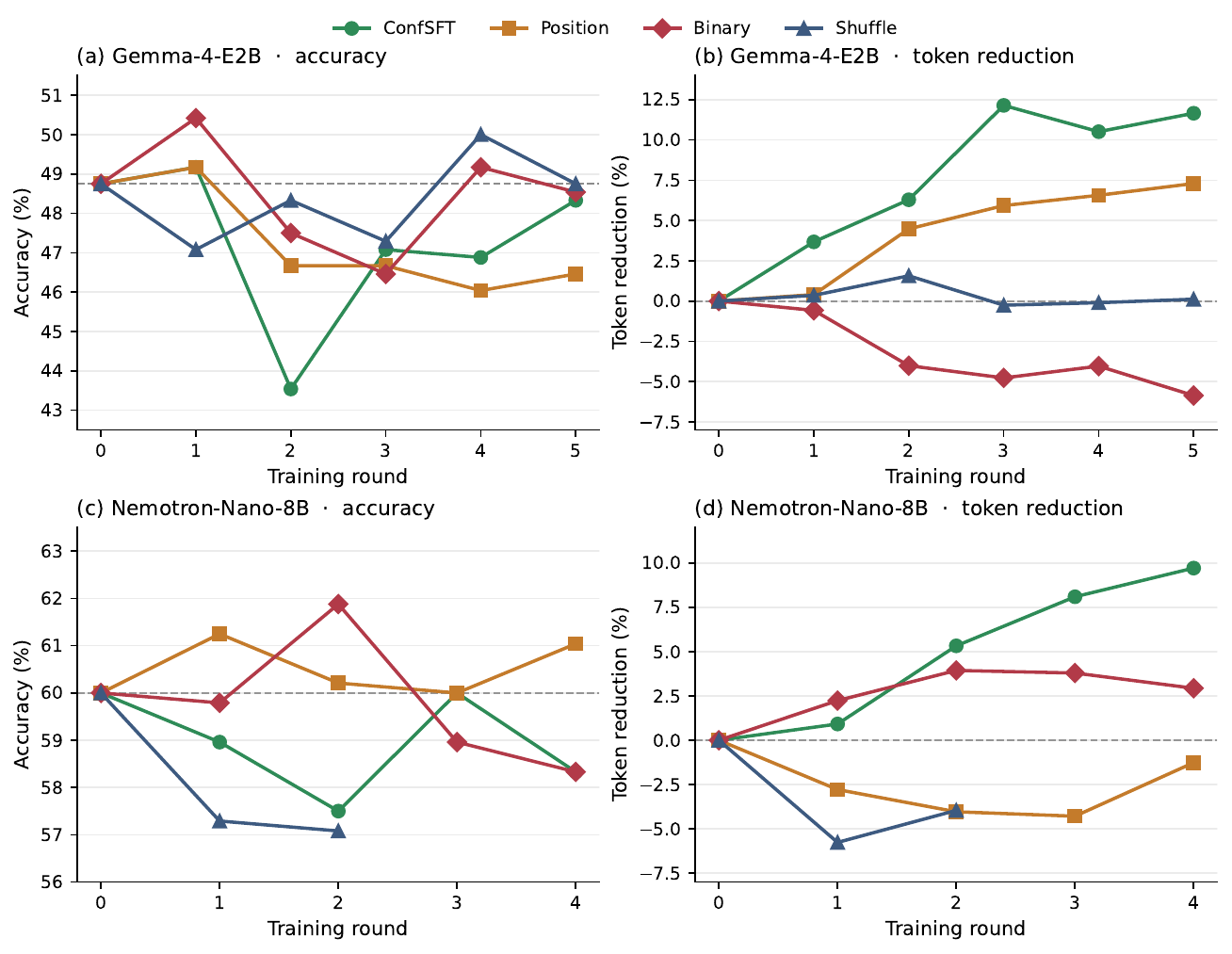}
  \caption{Accuracy (left) and token reduction (right) over training rounds
  for the four supervision signals in Table~\ref{tab:ablations}, AIME~2024
  validation.
  Dashed lines mark the pre-SFT baseline.}
  \label{fig:ablation-rounds}
\end{figure}

Figure~\ref{fig:ablation-rounds} tracks the four supervision variants of
Table~\ref{tab:ablations} on AIME~2024 validation ($30\times 16$) over training rounds.
Left panels show accuracy; right panels show token reduction relative to
the pre-SFT baseline (positive is shorter).
On both models, ConfSFT is the only variant whose token reduction grows
steadily while accuracy remains near the base model.

\section{Baselines}
\label{app:baselines}

\subsection{DEER}
\label{app:deer}

We apply the early-exit rule of \citet{deer} with confidence threshold
$0.95$ (following their setup) to stored reasoning traces and grade its output using the same
benchmark-specific procedures described above. Decision points and the
trial-answer probe match Appendices~\ref{app:prompts:anchors}
and~\ref{app:prompts:probe}. DEER exits at the first decision point with
$c_t\ge 0.95$ and returns the corresponding trial answer. If no decision
point crosses the threshold, it returns the original completion.

The reported token count includes the reasoning prefix through the exit and
the trial-answer tokens generated at each visited decision point. The fixed cue $q$ is not included in the token count.

On LiveCodeBench and HumanEval, the trial prompt replaces the boxed cue of
Appendix~\ref{app:prompts:probe}:
\begin{verbatim}
\n```python\n
\end{verbatim}
Following the original DEER setup for coding benchmarks, confidence is computed using at most the first $50$ trial-answer tokens.

We additionally evaluate a more conservative threshold of $0.98$ in Table~\ref{tab:deer-thr}. Increasing the threshold reduces premature exits and improves accuracy, but correspondingly yields smaller token savings. The qualitative conclusion remains unchanged: DEER's accuracy--efficiency behavior is highly sensitive to the stopping threshold and varies substantially across tasks, whereas ConfSFT achieves more consistent reductions without an inference-time stopping rule.

\begin{table}[t]
  \centering
  \caption{\textbf{DEER confidence thresholds.} Early exit at $0.95$ and $0.98$ thresholds.}
  \label{tab:deer-thr}
  \vspace{2pt}
  \setlength{\tabcolsep}{2.2pt}
  \renewcommand{\arraystretch}{1.12}
  \arrayrulecolor{tablerule}
  \scriptsize
  \newcommand{\hacc}{\multicolumn{1}{c}{Acc$\uparrow$}}
  \newcommand{\htok}{\multicolumn{1}{c}{Avg.\ Tokens$\downarrow$}}
  \newcommand{\hsav}{\multicolumn{1}{c}{Token Red.}}
  \newcommand{\haccG}{\multicolumn{1}{!{\color{tablerule}\vrule}c}{Acc$\uparrow$}}
  \newcommand{\dom}[2]{\multicolumn{#1}{c!{\color{tablerule}\vrule}}{\textsc{#2}}}
  \newcommand{\domend}[2]{\multicolumn{#1}{c}{\textsc{#2}}}
  \newcommand{\bn}[1]{\multicolumn{2}{c!{\color{tablerule}\vrule}}{#1}}
  \newcommand{\bnend}[1]{\multicolumn{2}{c}{#1}}
  \newcommand{\ts}[2]{#1\,\textcolor{savedgreen}{\tiny(#2\%)}}
  \newcommand{\sv}[1]{\textcolor{savedgreen}{#1\%}}
  \resizebox{\linewidth}{!}{%
  \begin{tabular}{@{}l !{\color{tablerule}\vrule}
      c@{\hspace{4pt}}c !{\color{tablerule}\vrule}
      c@{\hspace{4pt}}c !{\color{tablerule}\vrule}
      c@{\hspace{4pt}}c !{\color{tablerule}\vrule}
      c@{\hspace{4pt}}c !{\color{tablerule}\vrule}
      c@{\hspace{4pt}}c !{\color{tablerule}\vrule}
      c@{\hspace{4pt}}c@{}}
    \toprule
    \multirow{3}{*}{Method}
    & \dom{4}{Math} & \dom{2}{Science} & \dom{4}{Coding} & \domend{2}{Average} \\
    & \bn{AIME2025} & \bn{GSM8K} & \bn{GPQA-Diamond} & \bn{LiveCodeBench} & \bn{HumanEval} & \bnend{} \\
    \cmidrule(lr){2-3}\cmidrule(lr){4-5}\cmidrule(lr){6-7}\cmidrule(lr){8-9}\cmidrule(lr){10-11}\cmidrule(lr){12-13}
    & \hacc & \htok & \haccG & \htok
           & \haccG & \htok & \haccG & \htok
           & \haccG & \htok & \haccG & \hsav \\
    \midrule
    \multicolumn{13}{@{}l}{\textit{Nemotron-Nano-8B}} \\
    Base & 42.5 & 11,325 & 91.8 & 1,192 & 52.5 & 7,338 & 44.2 & 11,761 & 89.7 & 3,600 & 64.1 & \sv{0.0} \\
    DEER ($0.95$) & 39.8 & \ts{9,076}{-19.9} & 89.2 & \ts{765}{-35.8} & 49.8 & \ts{5,712}{-22.2} & 17.4 & \ts{1,731}{-85.3} & 47.1 & \ts{511}{-85.8} & 48.7 & \sv{-49.8} \\
    DEER ($0.98$) & 41.9 & \ts{9,585}{-15.4} & 90.4 & \ts{916}{-23.2} & 51.4 & \ts{6,494}{-11.5} & 28.1 & \ts{6,727}{-42.8} & 60.7 & \ts{1,005}{-72.1} & 54.5 & \sv{-33.0} \\
    \rowcolor{oursrow}
    \textbf{ConfSFT} & 45.6 & \ts{10,153}{-10.4} & 91.6 & \ts{1,010}{-15.3} & 52.8 & \ts{6,278}{-14.4} & 45.1 & \ts{11,151}{-5.2} & 90.8 & \ts{3,225}{-10.4} & 65.2 & \sv{-11.1} \\
    \midrule
    \multicolumn{13}{@{}l}{\textit{Gemma-4-E2B}} \\
    Base & 35.0 & 7,284 & 91.2 & 956 & 42.6 & 3,294 & 42.1 & 7,543 & 93.2 & 2,210 & 60.8 & \sv{0.0} \\
    DEER ($0.95$) & 24.2 & \ts{4,924}{-32.4} & 90.4 & \ts{912}{-4.6} & 36.5 & \ts{2,432}{-26.2} & 31.3 & \ts{5,333}{-29.3} & 82.9 & \ts{1,905}{-13.8} & 53.1 & \sv{-21.3} \\
    DEER ($0.98$) & 29.6 & \ts{5,814}{-20.2} & 90.7 & \ts{925}{-3.2} & 38.7 & \ts{2,739}{-16.8} & 38.9 & \ts{6,910}{-8.4} & 89.9 & \ts{2,136}{-3.4} & 57.6 & \sv{-10.4} \\
    \rowcolor{oursrow}
    \textbf{ConfSFT} & 32.1 & \ts{5,987}{-17.8} & 91.4 & \ts{938}{-1.9} & 40.8 & \ts{2,968}{-9.9} & 42.4 & \ts{6,403}{-15.1} & 93.8 & \ts{2,059}{-6.8} & 60.1 & \sv{-10.3} \\
    \midrule
    \multicolumn{13}{@{}l}{\textit{Qwen3-4B}} \\
    Base & 59.2 & 13,268 & 94.8 & 2,292 & 53.0 & 9,075 & 46.8 & 14,954 & 93.3 & 3,495 & 69.4 & \sv{0.0} \\
    DEER ($0.95$) & 40.4 & \ts{9,086}{-31.5} & 84.8 & \ts{580}{-74.7} & 47.1 & \ts{3,621}{-60.1} & 25.8 & \ts{10,374}{-30.6} & 50.8 & \ts{756}{-78.4} & 49.8 & \sv{-55.0} \\
    DEER ($0.98$) & 42.9 & \ts{9,289}{-30.0} & 87.4 & \ts{689}{-69.9} & 50.3 & \ts{4,573}{-49.6} & 39.2 & \ts{13,986}{-6.5} & 63.5 & \ts{1,140}{-67.4} & 56.7 & \sv{-44.7} \\
    \rowcolor{oursrow}
    \textbf{ConfSFT} & 58.8 & \ts{11,365}{-14.3} & 94.5 & \ts{1,710}{-25.4} & 55.8 & \ts{6,838}{-24.6} & 45.0 & \ts{12,122}{-18.9} & 93.3 & \ts{3,054}{-12.6} & 69.5 & \sv{-19.2} \\
    \midrule
    \multicolumn{13}{@{}l}{\textit{gpt-oss-20b}} \\
    Base & 52.7 & 5,802 & 94.0 & 482 & 61.0 & 3,732 & 53.5 & 5,034 & 95.9 & 917 & 71.4 & \sv{0.0} \\
    DEER ($0.95$) & 52.7 & \ts{5,807}{+0.1} & 94.0 & \ts{482}{+0.1} & 61.0 & \ts{3,753}{+0.6} & 51.1 & \ts{4,824}{-4.2} & 90.6 & \ts{832}{-9.2} & 69.9 & \sv{-2.5} \\
    DEER ($0.98$) & 52.7 & \ts{5,833}{+0.5} & 94.0 & \ts{483}{+0.2} & 61.0 & \ts{3,756}{+0.7} & 53.1 & \ts{5,267}{+4.6} & 94.4 & \ts{908}{-1.0} & 71.0 & \sv{+1.0} \\
    \rowcolor{oursrow}
    \textbf{ConfSFT} & 53.3 & \ts{5,235}{-9.8} & 94.1 & \ts{437}{-9.2} & 61.0 & \ts{3,112}{-16.6} & 50.7 & \ts{4,393}{-12.7} & 95.0 & \ts{850}{-7.2} & 70.8 & \sv{-11.1} \\
    \bottomrule
  \end{tabular}%
  }
\end{table}

\subsection{A\&Z}

We train Nemotron-8B and Gemma-4-E2B using the GRPO with the length penalty
from \cite{arora2025traininglanguagemodelsreason} on a random subset of size
10k from the DeepMath-103K dataset \cite{deepmath} for training.
We use the Verl framework for training, with FSDP and colocated vLLM rollout,
and TP=8 on $8{\times}$H100, using number of rollouts $K=16$ for Nemotron and
$K=8$ for Gemma-4, with batch size 64, learning rate $1\times10^{-6}$, KL loss
coefficient $0.04$, and no normalization of advantages by standard deviation in
GRPO. To fit the model into a H100 node, we use maximum response length of
$12{,}288$ for Nemotron and $9{,}216$ for Gemma-4, with GPU memory utilization
of $0.4$ and $0.3$ respectively. This is because Gemma-4 has a much larger
vocabulary size of ${\sim}262$k, which causes out-of-memory errors if the
maximum response length is too high. 

We trained our RL length-penalty models using a DeepMath-103K subset rather than the AIME. We use $\alpha=0.2$ for both Nemotron-8B and Gemma-4-E2B. Smaller values such as $\alpha=0.05$ yielded only minor efficiency gains, while larger values such as $\alpha=0.4$ caused accuracy degradation.

\begin{figure}[t]
  \centering
  \begin{minipage}[t]{0.49\linewidth}
    \centering
    \includegraphics[width=\linewidth]{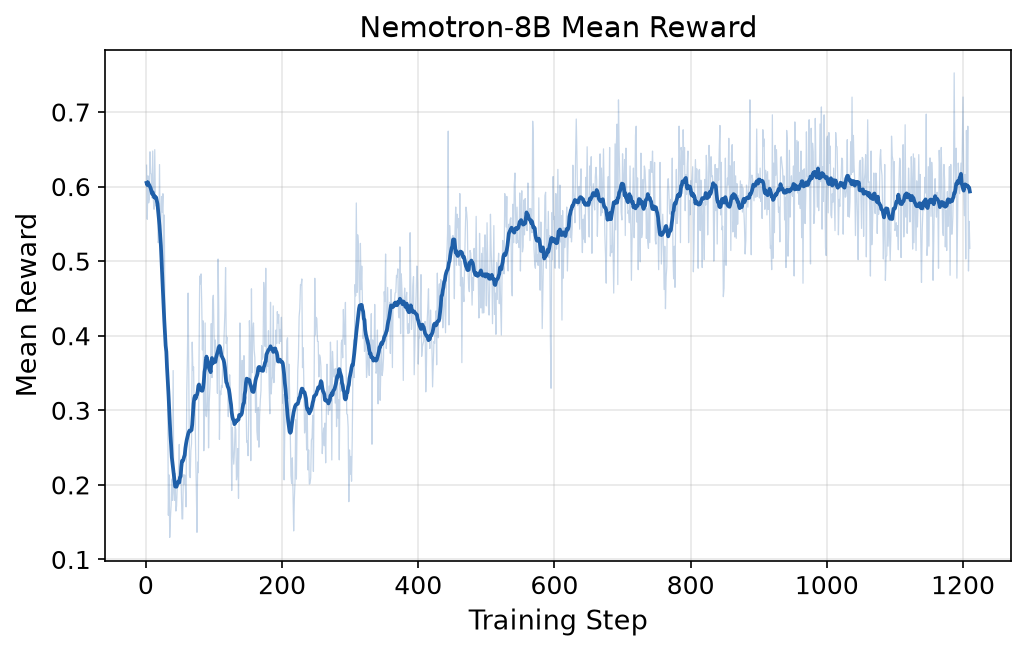}\\[2pt]
    {\small (a) Nemotron-8B reward.}
  \end{minipage}\hfill
  \begin{minipage}[t]{0.49\linewidth}
    \centering
    \includegraphics[width=\linewidth]{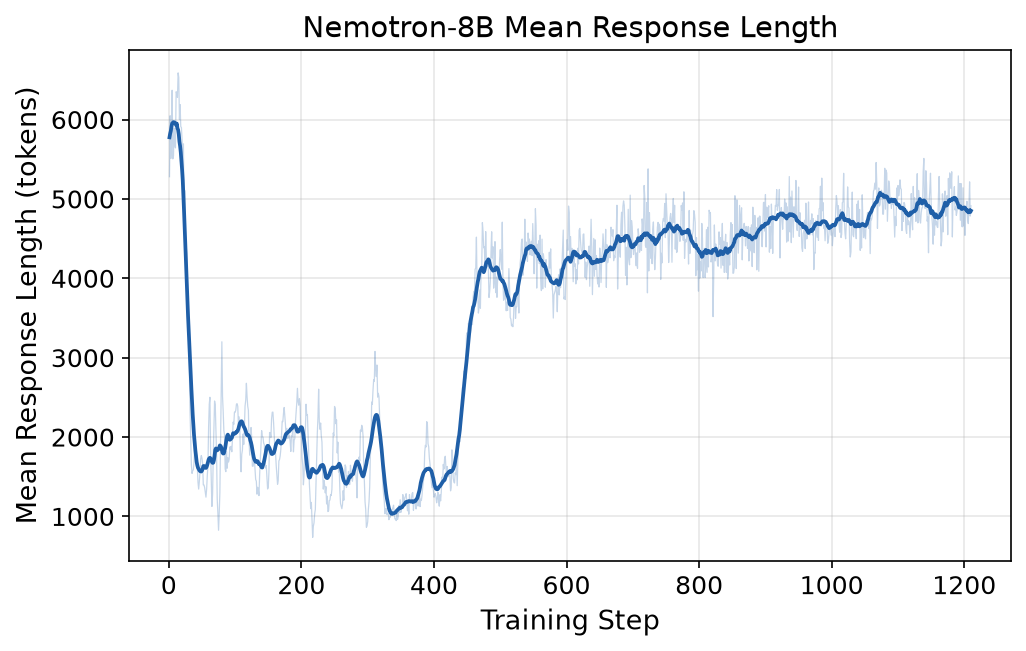}\\[2pt]
    {\small (b) Nemotron-8B mean response length.}
  \end{minipage}\\[6pt]
  \begin{minipage}[t]{0.49\linewidth}
    \centering
    \includegraphics[width=\linewidth]{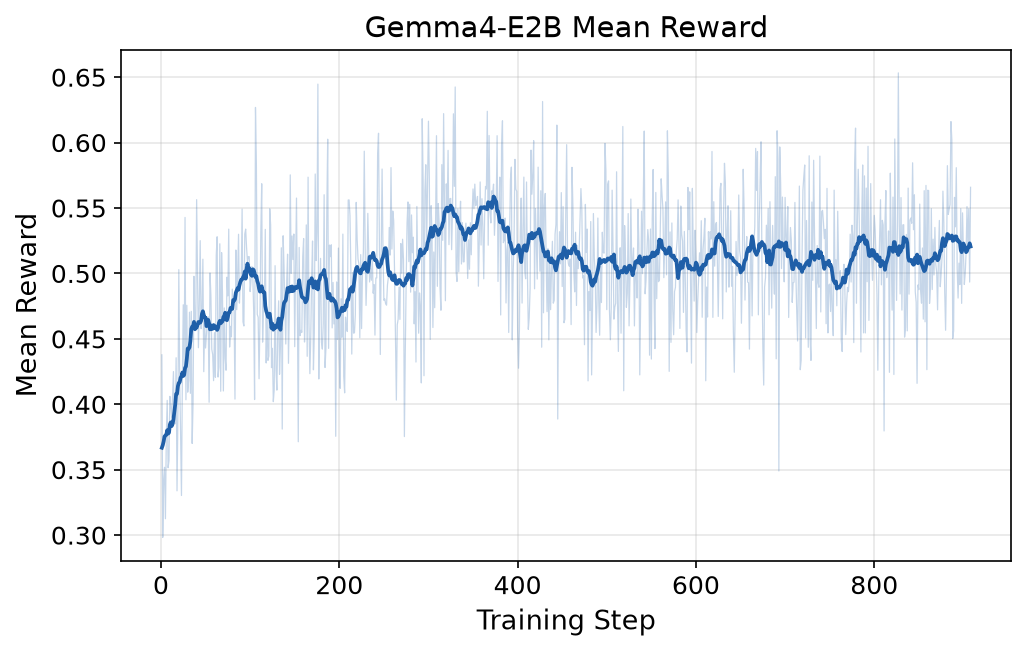}\\[2pt]
    {\small (c) Gemma-4-E2B reward.}
  \end{minipage}\hfill
  \begin{minipage}[t]{0.49\linewidth}
    \centering
    \includegraphics[width=\linewidth]{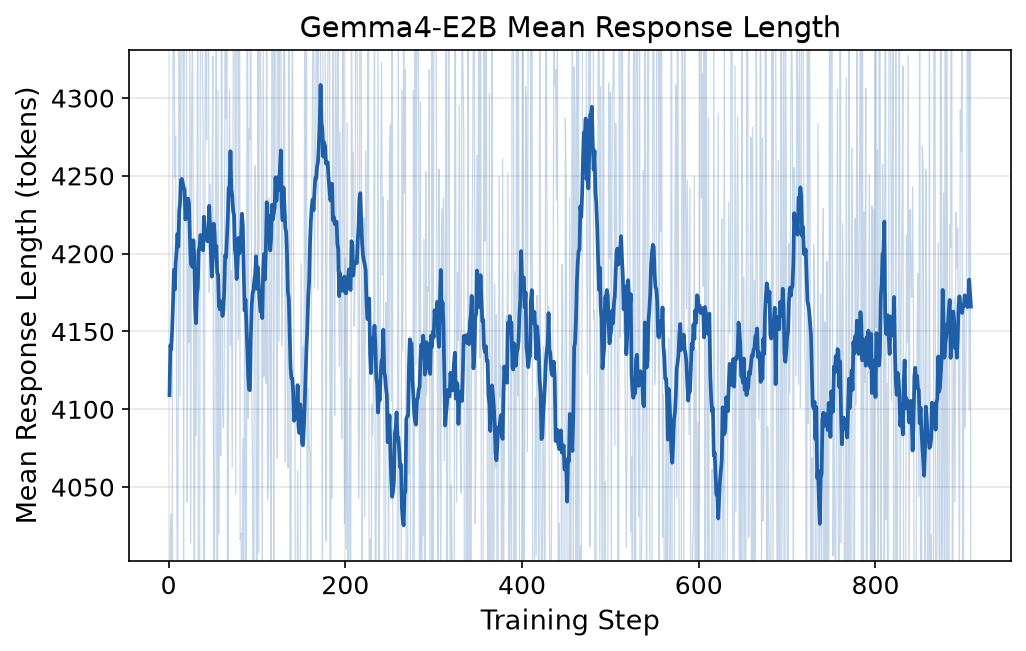}\\[2pt]
    {\small (d) Gemma-4-E2B mean response length.}
  \end{minipage}
  \caption{\textbf{Length-penalty GRPO training curves.}}
  \label{fig:rl-curves}
\end{figure}

We provide the A\&Z training curves in Figure~\ref{fig:rl-curves}. Compared with ConfSFT, this baseline uses a substantially larger training set (10K problems, roughly $15\times$ larger) and requires many more optimization steps. We evaluate multiple checkpoints from different training steps on AIME2025 and report the checkpoint with the greatest token reduction and accuracy close to the base model. Despite this additional training and checkpoint selection, A\&Z yields smaller token reductions than ConfSFT on both Nemotron and Gemma in Table~\ref{tab:main}.

The RL optimization is also noticeably less stable. On Nemotron, response length collapses early in training together with reward, and both require several hundred steps to recover. Gemma is more stable, but its response length decreases only modestly over training. We also find the method sensitive to the length-penalty coefficient: small values provide little efficiency gain, while larger values substantially degrade performance.

\subsection{On-Policy SFT}
We evaluate On-Policy-SFT~\citep{onpolicysft} as an additional baseline training method. We utilize their released code, which is a modification of the Verl framework, with some minor changes for version compatibility and adding support for additional models. For the Qwen3-4B, Nemotron, and Gemma-4 models, we use AIME 2000-2023 as a training dataset, with $N = 2$ rollouts per prompt per epoch, a training batch size of 16, a maximum response length of 16384. As a validation set, we use AIME 2024 with 16 rollouts per problem and a maximum respone length of 16384. For the Qwen and Gemma models, training is performed on 16$\times$RTXA5000 GPUs, while for Nemotron it is performed on 4$\times$L40S GPUs; we use tenor parallelism TP=2 in all cases.

For all three models, we perform a hyperparameter grid search over learning rate (ranging from $1\times 10^{-6}$ to $5\times 10^{-6}$). For all models, we use the recommended settings from each model's release for validation rollout generation; however, we set the generation training temperature at 1.0, following the guidelines in \citet{onpolicysft}. For each model, we run a training loop for each learning rate, saving a checkpoint after every 10 batches up to 2 epochs; among these checkpoints, we choose the one that provides the best accuracy-efficiency tradeoff on our validation set.


\section{Schoenfeld’s Anatomy}
\label{app:schoenfeld}
Schoenfeld's Episode Theory~\citep{schoenfeld2014mathematical} is a framework originally designed for analyzing the cognitive states of human during mathematical problem solving. \citet{Schoenfeld_emnlp, Schoenfeld} adapt the theory for studying the thinking patterns of large reasoning models. By decomposing the reasoning process into 8 categories of cognitive episodes, namely \textit{Read}, \textit{Analyze}, \textit{Plan}, \textit{Implement}, \textit{Explore}, \textit{Verify}, \textit{Answer}, and \textit{Monitor}, the adapted framework provides a comprehensive characterization of reasoning in terms of structures and dynamics.

We follow their sentence-level annotation setting and train a classifier so that the same taxonomy can be applied to every trace in our efficiency comparison. Supervision comes from the sentence-level episode labels on the open-source traces of \citet{Schoenfeld}, covering $15$ models. We draw a uniform subsample of $120{,}000$ sentences with seed $42$, then hold out a stratified $15\%$ test set, leaving $102{,}000$ training sentences and $18{,}000$ test sentences. As in the original annotation guide, the category of a sentence depends on both the sentence and its immediate context. Each training example therefore uses the previous sentence and the current sentence as input, and the label is one of the eight episode categories. The first sentence of a trace has an empty previous sentence.

We fine-tune \texttt{google-bert/bert-base-uncased}~\citep{devlin2019bert} with an eight-way classification head. Inputs are truncated at $160$ tokens. Training uses AdamW with learning rate $2\times 10^{-5}$ and $\epsilon=10^{-8}$, a linear decay schedule, and a warmup covering $6\%$ of the optimizer steps. Gradients are clipped to norm $1.0$, and the batch size is $32$. We train for three epochs. On the $18{,}000$ held-out sentences the final checkpoint reaches Cohen's $\kappa=0.754$, indicating strong agreement with the reference labels. 

At annotation time we segment each completion with the same sentence splitter used to build the training labels, and classify every sentence from its preceding sentence together with itself. Figure~\ref{fig:schoenfeld} compares each trained policy with the base policy it starts from. Both sides use the same AIME 2025 problems and the same 16 sampled completions per problem. We use a public checkpoint when one is available, and our reproduction otherwise. The share of episode $k$ in a completion is the fraction of its reasoning tokens assigned to sentences labeled $k$. For a checkpoint, $p_k$ is the average of this share over its completions. The heatmap reports the signed change $100\bigl(p_k^{\mathrm{after}}-p_k^{\mathrm{base}}\bigr)$ in percentage points. The bar for each method is the total variation
\begin{equation}
D_{\mathrm{TV}}
=
\frac{1}{2}\sum_{k=1}^{8}
\left|
p_k^{\mathrm{after}}-p_k^{\mathrm{base}}
\right|,
\end{equation}
the fraction of reasoning-token share reallocated across the eight episodes.

ConfSFT largely preserves the reasoning composition of the base policy it starts from. Its episode shares change only slightly, including on \emph{Implement}, \emph{Explore}, and \emph{Verify}, so the shorter traces still allocate computation across episodes in nearly the same way. The same comparison shows why this is specific to ConfSFT: L1-Max, ThinkPrune, and On-Policy SFT substantially redistribute computation across those episodes. ConfSFT reduces the amount of reasoning while keeping that allocation intact.


\end{document}